\def\year{2021}\relax
\documentclass[letterpaper]{article} 
\usepackage{aaai21}  
\usepackage{times}  
\usepackage{helvet} 
\usepackage{courier}  
\usepackage[hyphens]{url}  
\usepackage{graphicx} 
\usepackage{natbib}  
\usepackage{caption} 
\usepackage[ruled,vlined]{algorithm2e}

\usepackage{times}
\usepackage{latexsym}

\usepackage{url}
\usepackage[T1]{fontenc}
\usepackage{multirow}
\usepackage{tikz}
\usepackage{makecell}

\usepackage[inline]{enumitem}
\usepackage{bbm}

\usepackage{amsmath,amsfonts,bm}

\def\eqref#1{equation~\ref{#1}}

\def\1{\bm{1}}

\def\vc{{\bm{c}}}

\def\ve{{\bm{e}}}

\def\vs{{\bm{s}}}

\def\vw{{\bm{w}}}

\DeclareMathAlphabet{\mathsfit}{\encodingdefault}{\sfdefault}{m}{sl}
\SetMathAlphabet{\mathsfit}{bold}{\encodingdefault}{\sfdefault}{bx}{n}

\DeclareMathOperator*{\argmax}{arg\,max}
\DeclareMathOperator*{\argmin}{arg\,min}

\usepackage{hyperref} 

\def\lessgreat#1{\textless {#1}\textgreater{}}

\newcommand{\refexpsetup}{\ref{sec:exp_setup}}
\newcommand{\refsum}{\ref{sec:sum}}
\newcommand{\reftbvis}{\ref{tb:visual_example}}
\newcommand{\reftbphrase}{\ref{tb:phrase_sim}}
\newcommand{\reftbsts}{\ref{tb:STS}}
\newcommand{\reftbsum}{\ref{tb:sum}}
\newcommand{\refeqobj}{\eqref{eq:obj_func} }
\newcommand{\refeqer}{\eqref{eq:er} }
\newcommand{\refeqerdot}{\eqref{eq:er}.}
\newcommand{\refeqatt}{\eqref{eq:attention} }

\newcommand{\refappsentsim}{Appendix \ref{sec:sent_sim_more}}
\newcommand{\refapphyper}{Appendix \ref{sec:hyper}}
\newcommand{\refappexample}{Appendix \ref{sec:more_examples}}
\newcommand{\refappalgo}{Algorithm \ref{algo:NNSC} in the appendix}

\definecolor{c00}{rgb}{1 0.00 0.00}
\definecolor{c01}{rgb}{1 0.01 0.01}
\definecolor{c02}{rgb}{1 0.02 0.02}
\definecolor{c03}{rgb}{1 0.03 0.03}
\definecolor{c04}{rgb}{1 0.04 0.04}
\definecolor{c05}{rgb}{1 0.05 0.05}
\definecolor{c06}{rgb}{1 0.06 0.06}
\definecolor{c07}{rgb}{1 0.07 0.07}
\definecolor{c08}{rgb}{1 0.08 0.08}
\definecolor{c09}{rgb}{1 0.09 0.09}
\definecolor{c10}{rgb}{1 0.10 0.10}
\definecolor{c11}{rgb}{1 0.11 0.11}
\definecolor{c12}{rgb}{1 0.12 0.12}
\definecolor{c13}{rgb}{1 0.13 0.13}
\definecolor{c14}{rgb}{1 0.14 0.14}
\definecolor{c15}{rgb}{1 0.15 0.15}
\definecolor{c16}{rgb}{1 0.16 0.16}
\definecolor{c17}{rgb}{1 0.17 0.17}
\definecolor{c18}{rgb}{1 0.18 0.18}
\definecolor{c19}{rgb}{1 0.19 0.19}
\definecolor{c20}{rgb}{1 0.20 0.20}
\definecolor{c21}{rgb}{1 0.21 0.21}
\definecolor{c22}{rgb}{1 0.22 0.22}
\definecolor{c23}{rgb}{1 0.23 0.23}
\definecolor{c24}{rgb}{1 0.24 0.24}
\definecolor{c25}{rgb}{1 0.25 0.25}
\definecolor{c26}{rgb}{1 0.26 0.26}
\definecolor{c27}{rgb}{1 0.27 0.27}
\definecolor{c28}{rgb}{1 0.28 0.28}
\definecolor{c29}{rgb}{1 0.29 0.29}
\definecolor{c30}{rgb}{1 0.30 0.30}
\definecolor{c31}{rgb}{1 0.31 0.31}
\definecolor{c32}{rgb}{1 0.32 0.32}
\definecolor{c33}{rgb}{1 0.33 0.33}
\definecolor{c34}{rgb}{1 0.34 0.34}
\definecolor{c35}{rgb}{1 0.35 0.35}
\definecolor{c36}{rgb}{1 0.36 0.36}
\definecolor{c37}{rgb}{1 0.37 0.37}
\definecolor{c38}{rgb}{1 0.38 0.38}
\definecolor{c39}{rgb}{1 0.39 0.39}
\definecolor{c40}{rgb}{1 0.40 0.40}
\definecolor{c41}{rgb}{1 0.41 0.41}
\definecolor{c42}{rgb}{1 0.42 0.42}
\definecolor{c43}{rgb}{1 0.43 0.43}
\definecolor{c44}{rgb}{1 0.44 0.44}
\definecolor{c45}{rgb}{1 0.45 0.45}
\definecolor{c46}{rgb}{1 0.46 0.46}
\definecolor{c47}{rgb}{1 0.47 0.47}
\definecolor{c48}{rgb}{1 0.48 0.48}
\definecolor{c49}{rgb}{1 0.49 0.49}
\definecolor{c50}{rgb}{1 0.50 0.50}
\definecolor{c51}{rgb}{1 0.51 0.51}
\definecolor{c52}{rgb}{1 0.52 0.52}
\definecolor{c53}{rgb}{1 0.53 0.53}
\definecolor{c54}{rgb}{1 0.54 0.54}
\definecolor{c55}{rgb}{1 0.55 0.55}
\definecolor{c56}{rgb}{1 0.56 0.56}
\definecolor{c57}{rgb}{1 0.57 0.57}
\definecolor{c58}{rgb}{1 0.58 0.58}
\definecolor{c59}{rgb}{1 0.59 0.59}
\definecolor{c60}{rgb}{1 0.60 0.60}
\definecolor{c61}{rgb}{1 0.61 0.61}
\definecolor{c62}{rgb}{1 0.62 0.62}
\definecolor{c63}{rgb}{1 0.63 0.63}
\definecolor{c64}{rgb}{1 0.64 0.64}
\definecolor{c65}{rgb}{1 0.65 0.65}
\definecolor{c66}{rgb}{1 0.66 0.66}
\definecolor{c67}{rgb}{1 0.67 0.67}
\definecolor{c68}{rgb}{1 0.68 0.68}
\definecolor{c69}{rgb}{1 0.69 0.69}
\definecolor{c70}{rgb}{1 0.70 0.70}
\definecolor{c71}{rgb}{1 0.71 0.71}
\definecolor{c72}{rgb}{1 0.72 0.72}
\definecolor{c73}{rgb}{1 0.73 0.73}
\definecolor{c74}{rgb}{1 0.74 0.74}
\definecolor{c75}{rgb}{1 0.75 0.75}
\definecolor{c76}{rgb}{1 0.76 0.76}
\definecolor{c77}{rgb}{1 0.77 0.77}
\definecolor{c78}{rgb}{1 0.78 0.78}
\definecolor{c79}{rgb}{1 0.79 0.79}
\definecolor{c80}{rgb}{1 0.80 0.80}
\definecolor{c81}{rgb}{1 0.81 0.81}
\definecolor{c82}{rgb}{1 0.82 0.82}
\definecolor{c83}{rgb}{1 0.83 0.83}
\definecolor{c84}{rgb}{1 0.84 0.84}
\definecolor{c85}{rgb}{1 0.85 0.85}
\definecolor{c86}{rgb}{1 0.86 0.86}
\definecolor{c87}{rgb}{1 0.87 0.87}
\definecolor{c88}{rgb}{1 0.88 0.88}
\definecolor{c89}{rgb}{1 0.89 0.89}
\definecolor{c90}{rgb}{1 0.90 0.90}
\definecolor{c91}{rgb}{1 0.91 0.91}
\definecolor{c92}{rgb}{1 0.92 0.92}
\definecolor{c93}{rgb}{1 0.93 0.93}
\definecolor{c94}{rgb}{1 0.94 0.94}
\definecolor{c95}{rgb}{1 0.95 0.95}
\definecolor{c96}{rgb}{1 0.96 0.96}
\definecolor{c97}{rgb}{1 0.97 0.97}
\definecolor{c98}{rgb}{1 0.98 0.98}
\definecolor{c99}{rgb}{1 0.99 0.99}
\definecolor{c100}{rgb}{1 1.00 1.00}

\title{Extending Multi-Sense Word Embedding to Phrases and Sentences \\ for Unsupervised Semantic Applications}

\author{
    Haw-Shiuan Chang, Amol Agrawal, Andrew McCallum \\   
}
\affiliations{

    CICS, University of Massachusetts Amherst \\
    {\{hschang,amolagrawal,mccallum\}@cs.umass.edu}

}

\begin{document}

\maketitle

\begin{abstract}
Most unsupervised NLP models represent each word with a single point or single region in semantic space, while the existing multi-sense word embeddings cannot represent longer word sequences like phrases or sentences. We propose a novel embedding method for a text sequence (a phrase or a sentence) where each sequence is represented by a distinct set of multi-mode codebook embeddings to capture different semantic facets of its meaning. The codebook embeddings can be viewed as the cluster centers which summarize the distribution of possibly co-occurring words in a pre-trained word embedding space. We introduce an end-to-end trainable neural model that directly predicts the set of cluster centers from the input text sequence during test time. Our experiments show that the per-sentence codebook embeddings significantly improve the performances in unsupervised sentence similarity and extractive summarization benchmarks. In phrase similarity experiments, we discover that the multi-facet embeddings provide an interpretable semantic representation but do not outperform the single-facet baseline.
\end{abstract}

\section{Introduction}
\label{sec:intro}



Collecting manually labeled data is an expensive and tedious process for new or low-resource NLP applications. 
Many of these applications require the text similarity measurement based on the text representation learned from 
the raw text without any supervision. 
Examples of the representation include word embedding like Word2Vec~\citep{word2vec} or GloVe~\citep{glove}, sentence embeddings like skip-thoughts~\citep{kiros2015skip}, contextualized word embedding like ELMo~\citep{ELMo} and BERT~\citep{BERT} without fine-tuning. 

The existing work often represents a word sequence (e.g., a sentence or a phrase) as a single embedding. However, when squeezing all the information into a single embedding (e.g., by averaging the word embeddings or using CLS embedding in BERT), the representation might lose some important information of different facets in the sequence.






\begin{figure}[t!]
\begin{center}
\includegraphics[width=1\linewidth]{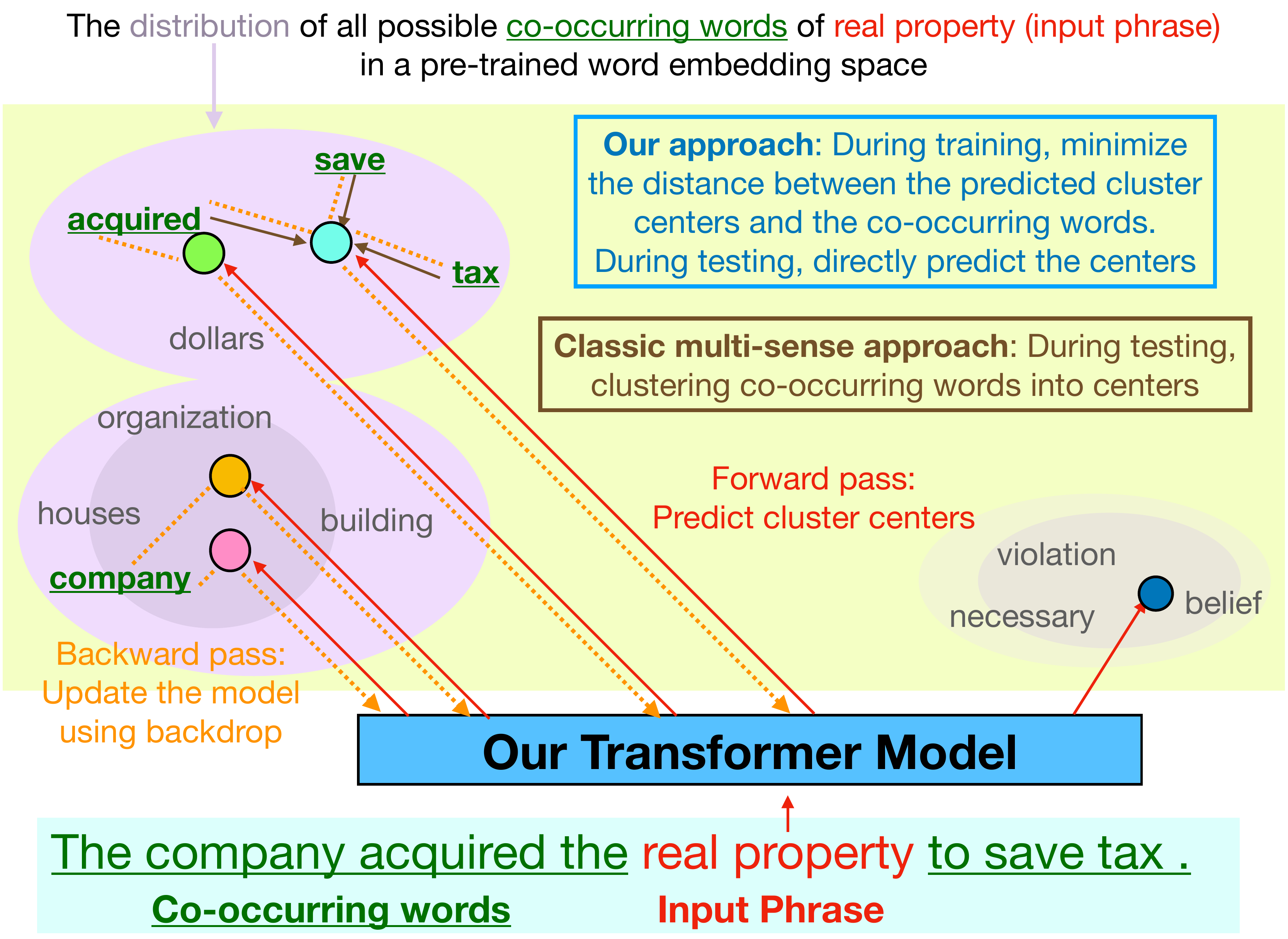}
\end{center}
\caption{The input phrase \emph{real property} is represented by $K=5$ cluster centers. The previous work discovers the multiple senses by 
clustering the embedding of observed co-occurring words.
Instead, our compositional model learns to predict the embeddings of cluster centers from the sequence of words in the input phrase so as to reconstruct the (unseen) co-occurring distribution well.}
\label{fig:first_page}
\end{figure}









Inspired by the multi-sense word embeddings~\citep{lau2012word,NeelakantanSPM14, athiwaratkun2017multimodal, singh2020context}, we propose a multi-facet representation that characterizes a phrase or a sentence as a fixed number of embeddings, where each embedding is a clustering center of the words co-occurring with the input word sequence. 

In this work, a facet refers to a mode of the co-occurring word distribution, which might be multimodal. For example, the multi-facet representation of \emph{real property} is illustrated in Figure~\ref{fig:first_page}. Real property can be observed in legal documents where it usually means real estate, while real property can also mean a true characteristic in philosophic discussions. The previous unsupervised multi-sense embeddings discover those senses by clustering the observed neighboring words (e.g., \emph{acquired}, \emph{save}, and \emph{tax}) and an important facet, a mode with high probability, could be represented by several close cluster centers. Notice that the approaches need to solve a distinct local clustering problem for each phrase in contrast with the topic modeling like LDA~\citep{blei2003latent}, which clusters all the words in the corpus into a global set of topics. 







\begin{figure*}[t!]
\centering
\includegraphics[width=0.9\linewidth]{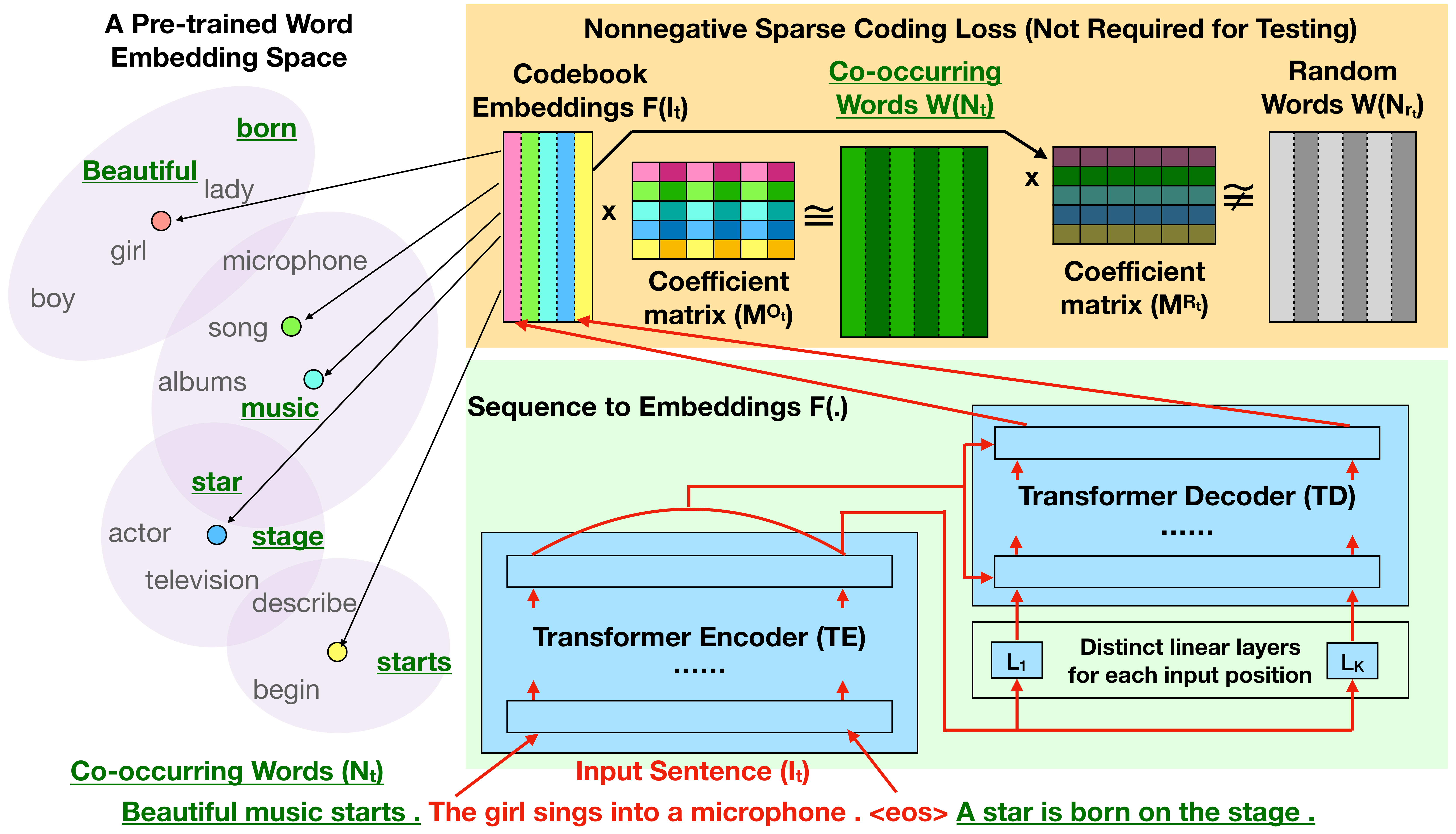}
\caption{Our model for sentence representation. We represent each sentence as multiple codebook embeddings (i.e., cluster centers) predicted by our sequence to embeddings model. Our loss encourages the model to generate codebook embeddings whose linear combination can well reconstruct the embeddings of co-occurring words (e.g., \emph{music}), while not able to reconstruct the negatively sampled words (i.e., the co-occurring words from other sentences).} 
\label{fig:architecture}
\end{figure*}

In addition to a phrase, we can also cluster the nearby words of a sentence which appears frequently in the corpus. The cluster centers usually correspond to important aspects rather than senses (see an example in Figure~\ref{fig:architecture}) because a sentence usually has multiple aspects but only one sense. However, extending the clustering-based multi-sense word embeddings to long sequences such as sentences is difficult in practice due to two efficiency challenges. First, there are usually many more unique phrases and sentences in a corpus than there are words, while the number of parameters for clustering-based approaches is $O(|V| \times |K| \times |E|)$, where $|V|$ is the number of unique sequences, $|K|$ is the number of clusters, and $|E|$ is the embedding dimensions. Estimating and storing such a large number of parameters takes time and space. More importantly, much more unique sequences imply much fewer co-occurring words to be clustered for each sequence, especially for sentences. An effective model needs to overcome this sample efficiency challenge (i.e., sparseness in the co-occurring statistics), but clustering approaches often have too many parameters to learn the compositional meaning of each sequence without overfitting.


Nevertheless, the sentences (or phrases) sharing multiple words often lead to similar cluster centers, so we should be able to solve these local clustering problems using much fewer parameters to circumvent the challenges. 
To achieve the goal, we develop a novel Transformer-based neural encoder and decoder. As shown in Figure~\ref{fig:first_page}, instead of clustering co-occurring words beside an input sequence at test time as in previous approaches, we learn a mapping between the input sequence (i.e., phrases or sentences) and the corresponding cluster centers during training so that we can directly predict those cluster centers using a single forward pass of the neural network for an arbitrary unseen input sequence during testing.

To train the neural model that predicts the clustering centers, we match the sequence of predicted cluster centers and the observed set of co-occurring word embeddings using a non-negative and sparse permutation matrix.
After the permutation matrix is estimated for each input sequence, the gradients are back-propagated to cluster centers (i.e.,  codebook embeddings) and to the weights of our neural model, which allows us to train the whole model end-to-end. 


In the experiments, we evaluate whether the proposed multi-facet embeddings could improve the similarity measurement between two sentences, between a sentence and a document (i.e., extractive summarization), and between phrases. The results demonstrate multi-facet embeddings significantly outperforms the classic single embedding baseline when the input sequence is a sentence. 

We also demonstrate several advantages of the proposed multi-facet embeddings over the (contextualized) embeddings of all the words in a sequence. First, we discover that our model tends to use more embeddings to represent an important facet or important words. This tendency provides an unsupervised estimation of word importance, which improves various similarity measurements between a sentence pair. Second, our model outputs a fixed number of facets by compressing long sentences and extending short sentences. In unsupervised extractive summarization, this capability prevents the scoring function from biasing toward longer or shorter sentences. Finally, in the phrase similarity experiments, our methods capture the compositional meaning (e.g., a \emph{hot dog} is a food) of a word sequence well and the quality of our similarity estimation is not sensitive to the choice of \emph{K}, the number of our codebook embeddings.







\subsection{Main Contributions}
\label{sec:contributions}
\setlist{nolistsep}
\begin{enumerate}
    \item As shown in Figure~\ref{fig:first_page},  
    we propose a novel framework that predicts the cluster centers of co-occurring word embeddings to overcomes the sparsity challenges in our self-supervised training signals.
    This allows us to extend the idea of clustering-based multi-sense embeddings to phrases or sentences.
    \item We propose a deep architecture that can effectively encode a sequence and decode a set of embeddings. We also propose non-negative sparse coding (NNSC) loss to train our neural encoder and decoder end-to-end.
    \item We demonstrate how the multi-facet embeddings could be used in unsupervised ways to improve the similarity between sentences/phrases, infer word importance in a sentence, extract important sentences in a document. In \refapphyper, we show that our model could provide asymmetric similarity measurement for hypernym detection. 
    \item We conduct comprehensive experiments in the main paper and appendix to show that multi-facet embedding is consistently better than classic single-facet embedding for modeling the co-occurring word distribution of sentences, while multi-facet phrase embeddings do not yield a clear advantage against the single embedding baseline, which supports the finding in \citet{dubossarsky2018coming}.

\end{enumerate}


\section{Method}
\label{sec:method}

In this section, we first formalize our training setup and next describe our objective function and neural architecture. Our approach is visually summarized in Figure~\ref{fig:architecture}. 

 


\subsection{Self-supervision Signal}
\label{sec:method_goal}

We express $t$th sequence of words in the corpus as $I_t = w_{x_t} ... w_{y_t}$\lessgreat{eos}, where $x_t$ and $y_t$ are the start and end position of the input sequence, respectively, and \lessgreat{eos} is the end of sequence symbol.

We assume neighboring words beside each input phrase or sentence are related to some facets of the sequence, so given $I_t$ as input, our training signal is to reconstruct a set of co-occurring words, 
$N_t = \{w_{x_t-d^t_1}, ... w_{x_t-1}, w_{y_t+1}, ... w_{y_t+d^t_2}\}$.\footnote{The self-supervised signal is a generalization of the loss for prediction-based word embedding like Word2Vec~\citep{word2vec}. They are the same when the input sequence length $|I_t|$ is 1.} In our experiments, we train our multi-facet sentence embeddings by setting $N_t$ as the set of all words in the previous and the next sentence, and train multi-facet phrase embeddings by setting a fixed window size $d^t_1=d^t_2=5$.


Since there are not many co-occurring words for a long sequence (none are observed for unseen testing sequences), the goal of our model is to predict the cluster centers of the words that could "possibly" occur beside the text sequence rather than the cluster centers of the actual occurring words in $N_t$ (e.g., the hidden co-occurring distribution instead of green and underlined words in Figure~\ref{fig:architecture}). The cluster centers of an unseen testing sequence are predictable because the model could learn from similar sequences and their co-occurring words in the training corpus.



To focus on the semantics rather than syntax, we view the co-occurring words as a set rather than a sequence as in skip-thoughts~\citep{kiros2015skip}. 
Notice that our model considers the word order information in the input sequence $I_t$, but ignores the order of the co-occurring words $N_t$.







\subsection{Non-negative Sparse Coding Loss}
\label{sec:method_loss}
In a pre-trained word embedding space, we predict the cluster centers of the co-occurring word embeddings. The embeddings of co-occurring words $N_t$ are arranged into a matrix $\bm{W(N_t)} = [ \underline{\vw}^t_j]_{j=1...{|N_t|}}$ with size $|E| \times |N_t|$, where $|E|$ is the dimension of pre-trained word embedding, and each of its columns $\underline{\vw}^t_j$ is a normalized word embedding whose 2-norm is 1. The normalization makes the cosine distance between two words become half of their squared Euclidean distance. 

Similarly, we denote the predicted cluster centers $\underline{\vc}^t_k$ of the input sequence $I_t$ as a $|E| \times K$ matrix $\bm{F(I_t)} = [ \underline{\vc}^t_k]_{k=1...K}$, where $\bm{F}$ is our neural network model and $K$ is the number of clusters. 
We fix the number of clusters $K$ to simplify the design of our prediction model and the unsupervised scoring functions used in the downstream tasks. When the number of modes in the (multimodal) co-occurring distribution is smaller than $K$, the model can output multiple cluster centers to represent a mode (e.g., the \emph{music} facet in Figure~\ref{fig:architecture} is represented by two close cluster centers).
As a result, the performances in our downstream applications are not sensitive to the setting of $K$ when $K$ is larger than the number of facets in most input word sequences.

The reconstruction loss of k-means clustering in the word embedding space 
can be written as $||\bm{F(I_t)}\bm{M} - \bm{W(N_t)}||^2 = \sum_j || (\sum_k \bm{M}_{k,j} \underline{\vc}^t_k)  - \underline{\vw}^t_j ||^2$, where $\bm{M}_{k,j}=1$ if the $j$th word belongs to the $k$ cluster and 0 otherwise. That is, $\bm{M}$ is a permutation matrix which matches the cluster centers and co-occurring words and allow the cluster centers to be predicted in an arbitrary order.

Non-negative sparse coding (NNSC)~\citep{hoyer2002non} relaxes the constraints by allowing the coefficient $\bm{M}_{k,j}$ to be a positive value but encouraging it to be 0. We adopt NNSC in this work because we observe that the neural network trained by NNSC loss generates more diverse topics than k-means loss does. We hypothesize that it is because the loss is smoother and easier to be optimized for a neural network. Using NNSC, we define our reconstruction error as

\vspace{-3mm}
\small
\begin{align}
& Er(\bm{F(I_t)},\bm{W(N_t)}) = ||\bm{F(I_t)}\bm{M^{O_t}} - \bm{W(N_t)}||^2 \nonumber \\ 
& s.t., \bm{M^{O_t}} = \argmin_{\bm{M}} ||\bm{F(I_t)}\bm{M} - \bm{W(N_t)}||^2 + \lambda || \bm{M} ||_1, \nonumber \\ 
& \forall k,j, \; 0 \leq \bm{M}_{k,j} \leq 1, 
\label{eq:er}
\end{align}
\normalsize
where $\lambda$ is a hyper-parameter controlling the sparsity of $\bm{M}$. We force the coefficient value $\bm{M}_{k,j}\leq 1$ to avoid the neural network learning to predict centers with small magnitudes which makes the optimal values of $\bm{M}_{k,j}$ large and unstable.


We adopt an alternating optimization strategy similar to the EM algorithm for k-means. At each iteration, our E-step estimates the permutation coefficient $\bm{M^{O_t}}$ after fixing our neural model, while our M-step treats $\bm{M^{O_t}}$ as constants to back-propagate the gradients of NNSC loss to our neural network.
A pseudo-code of our training procedure could be found in \refappalgo.
Estimating the permutation between the prediction and ground truth words is often computationally expensive~\citep{label_set}. Nevertheless, optimizing the proposed loss is efficient because 
for each training sequence $I_t$, $\bm{M^{O_t}}$ can be efficiently estimated using convex optimization (our implementation uses RMSprop~\citep{tieleman2012lecture}). Besides, we minimize the L2 distance, $||\bm{F(I_t)}\bm{M^{O_t}} - \bm{W(N_t)}||^2$, in a pre-trained embedding space as in~\citet{kumar2018mises,li2019efficient} rather than computing softmax. 

To prevent the neural network from predicting the same global topics regardless of the input, our loss function for $t$th sequence is defined as

\vspace{-3mm}
\footnotesize
\begin{align}
L_t(\bm{F}) = Er( \bm{F(I_t)} , \bm{W(N_t)} ) -   Er( \bm{F(I_t)} , \bm{W(N_{r_t}} )),
\label{eq:obj_func}
\end{align}
\normalsize
\noindent where $N_{r_t}$ is a set of co-occurring words of a randomly sampled sequence $I_{r_t}$. In our experiment, we use SGD to solve $\widehat{\bm{F}} = \argmin_{\bm{F}} \sum_t L_t(\bm{F})$. Our method could be viewed as a generalization of Word2Vec~\citep{word2vec} that can encode the compositional meaning of the words and decode multiple embeddings.

\subsection{Sequence to Embeddings}
\label{sec:method_neural}

Our neural network architecture is similar to Transformer-based sequence to sequence (seq2seq) model~\citep{vaswani2017attention}. We use the same encoder $TE(I_t)$, which transforms the input sequence into a contextualized embeddings

\vspace{-3mm}
\small
\begin{align}
[\underline{\ve}_{x_t} ... \underline{\ve}_{y_t} \underline{\ve}_{\text{\lessgreat{eos}}}] = TE( w_{x_t} ... w_{y_t}\text{\lessgreat{eos}} ),
\end{align}
\normalsize
where the goal of the encoder is to map the similar sentences, which are likely to have similar co-occurring word distribution, to similar contextualized embeddings.

Different from the typical seq2seq model~\citep{sutskever2014sequence, vaswani2017attention}, our decoder does not need to make discrete decisions because our outputs are a sequence of embeddings instead of words. This allows us to predict all the codebook embeddings in a single forward pass as in~\citet{lee2018set} without requiring an expensive softmax layer or auto-regressive decoding.\footnote{The decoder can also be viewed as another Transformer encoder which attends the output of the first encoder and models the dependency between predicted cluster centers.}


To make different codebook embeddings capture different facets, we pass the embeddings of \lessgreat{eos}, $\underline{\ve}_{\text{\lessgreat{eos}}}$, to different linear layers $L_k$ before becoming the input of the decoder $TD$. The decoder allows the input embeddings to attend each other to model the dependency among the facets and attend the contextualized word embeddings from the encoder, $\underline{\ve}_{x_t} ... \underline{\ve}_{y_t}\underline{\ve}_{\text{\lessgreat{eos}}}$, to copy the embeddings of some keywords in the word sequence as our facet embeddings more easily.
Specifically, the codebook embeddings


\vspace{-3mm}
\small
\begin{align}
\bm{F(I_t)} = TD(L_1(\underline{\ve}_{\text{\lessgreat{eos}}}) ... L_K(\underline{\ve}_{\text{\lessgreat{eos}}}), \underline{\ve}_{x_t} ... \underline{\ve}_{y_t}\underline{\ve}_{\text{\lessgreat{eos}}}).
\end{align}
\normalsize

We find that removing the attention on the $\underline{\ve}_{x_t} ... \underline{\ve}_{y_t}\underline{\ve}_{\text{\lessgreat{eos}}}$ significantly deteriorates our validation loss for sentence representation because there are often too many facets to be compressed into a single embedding. On the other hand, the encoder-decoder attention does not significantly change the performance of phrase representation, so we remove the connection (i.e., encoder and decoder have the same architecture) in models for phrase representation.
Notice that the framework is flexible. For example, we can encode the genre of the document containing the sentence if desired.






\section{Experiments}
\label{sec:exp}

Quantitatively evaluating the quality of our predicted cluster centers is difficult because the existing label data and metrics are built for global clustering. The previous multi-sense word embedding studies often show that multiple embeddings could improve the single word embedding in the unsupervised word similarity task to demonstrate its effectiveness. Thus, our goal of experiments is to discover when and how the multi-facet embeddings can improve the similarity measurement in various unsupervised semantic tasks upon the widely-used general-purpose representations, such as single embedding and (contextualized) word embeddings.




\subsection{Experiment Setup}
\label{sec:exp_setup}
Our models only require the raw corpus and sentence/phrase boundaries, so we will only compare them with other unsupervised alternatives that do not require any manual labels or multi-lingual resources such as PPDB~\citep{pavlick2015ppdb}. To simplify the comparison, we also omit the comparison with the methods using character-level information such as fastText~\citep{bojanowski2017enriching} or bigram information such as Sent2Vec~\citep{pagliardini2018unsupervised}.


It is hard to make a fair comparison with BERT~\citep{BERT}. Its masked language modeling loss is designed for downstream supervised tasks and 
preserves more syntax information which might be distractive in unsupervised semantic applications. Furthermore, BERT uses word piece tokenization while other models use word tokenization. Nevertheless, we still present the performances of the BERT Base model as a reference even though it is trained using more parameters, larger embedding size, larger corpus, and more computational resources compared with our models. Since we focus on unsupervised setting, we directly use the final hidden states of the BERT models without supervised fine-tuning in most of the comparisons. One exception is that we also report the performance of sentence-BERT~\citep{reimers2019sentence} in a low-resource setting.



Our model is trained on English Wikipedia 2016 while the stop words are removed from the set of co-occurring words. In the phrase experiments, we only consider noun phrases, and their boundaries are extracted by applying simple regular expression rules to POS tags before training. We use the cased version (840B) of GloVe embedding~\citep{glove} as the pre-trained word embedding space for our sentence representation and use the uncased version (42B) for phrase representation.\footnote{\footnotesize\url{nlp.stanford.edu/projects/glove/}} To control the effect of embedding size, we set the hidden state size in our transformers as the GloVe embedding size (300).

Limited by computational resources, we train all the models using one GPU (e.g., NVIDIA 1080 Ti) within a week. Because of the relatively small model size, we find that our models underfit the data after a week (i.e., the training loss is very close to the validation loss). 

\begin{table}[t!]

\scalebox{0.85}{
\begin{tabular}{|p{9.3cm}|}
\Xhline{5\arrayrulewidth}
\textbf{Input Phrase}: civil order \lessgreat{eos} \\ \Xhline{5\arrayrulewidth}
\textbf{Output Embedding (K = 1)}: \\ 
e1 \: | \space government 0.817 civil 0.762 citizens 0.748 \\ \hline
\textbf{Output Embeddings (K = 3)}: \\ 
e1 \: | \space initiatives 0.736 organizations 0.725 efforts 0.725 \\
e2 \: | \space army 0.815 troops 0.804 soldiers 0.786 \\
e3 \: | \space court 0.758 federal 0.757 judicial 0.736 \\ \hline
\multicolumn{1}{c}{}\\ \Xhline{5\arrayrulewidth}
\textbf{Input Sentence}: SMS messages are used in some countries as reminders of hospital appointments . \lessgreat{eos} \\ \Xhline{5\arrayrulewidth}
\textbf{Output Embedding (K = 1)}: \\  
 e1 \: | \space information 0.702, use 0.701, specific 0.700 \\ \hline
\textbf{Output Embeddings (K = 3)}: \\  
 e1 \: | \space can 0.769, possible 0.767, specific 0.767 \\
 e2 \: | \space hospital 0.857, medical 0.780, hospitals 0.739 \\
 e3 \: | \space SMS 0.791, Mobile 0.635, Messaging 0.631 \\ \hline
 \textbf{Output Embeddings (K = 10)}: \\
 e1 \: | \space can 0.854, should 0.834, either 0.821 \\
 e2 \: | \space hospital 0.886, medical 0.771, hospitals 0.745 \\
 e3 \: | \space services 0.768, service 0.749, web 0.722 \\
 e4 \: | \space  SMS 0.891, sms 0.745, messaging 0.686 \\
 e5 \: | \space messages 0.891, message 0.801, emails 0.679 \\
 e6 \: | \space systems 0.728, technologies 0.725, integrated 0.723 \\
 e7 \: | \space appointments 0.791, appointment 0.735, duties 0.613 \\
 e8 \: | \space confirmation 0.590, request 0.568, receipt 0.563 \\ 
 e9 \: | \space countries 0.855, nations 0.737, Europe 0.732 \\
 e10 | \space Implementation 0.613, Application 0.610, Programs 0.603\\ \hline 
\end{tabular}
}
\centering
\caption{Examples of the codebook embeddings predicted by our models with different $K$. The embedding in each row is visualized by the three words whose GloVe embeddings have the highest cosine similarities (also presented) with the codebook embedding.}
\label{tb:visual_example}
\end{table}

\subsection{Qualitative Evaluation}
The cluster centers predicted by our model are visualized in Table~\ref{tb:visual_example} (as using \emph{girl} and \emph{lady} to visualize the red cluster center in Figure~\ref{fig:architecture}). Some randomly chosen examples are also visualized in \refappexample.

The centers summarize the input sequence well and more codebook embeddings capture more fine-grained semantic facets of a phrase or a sentence. Furthermore, the embeddings capture the compositional meaning of words. For example, each word in the phrase \emph{civil order} does not mean \emph{initiatives}, \emph{army}, or \emph{court}, which are facets of the whole phrase. When the input is a sentence, we can see that the output embeddings are sometimes close to the embeddings of words in the input sentence, which explains why attending the contextualized word embeddings in our decoder 
could improve the quality of the output embeddings.


\subsection{Unsupervised Sentence Similarity}
\label{sec:STS}
We propose two ways to evaluate the multi-facet embeddings using sentence similarity tasks. 

\textbf{First way}: Since similar sentences should have similar word distribution in nearby sentences and thus similar codebook embeddings, the codebook embeddings of a query sentence $\bm{\widehat{F}_u(S^1_q)}$ should be able to well reconstruct the codebook embeddings of its similar sentence $\bm{\widehat{F}_u(S^2_q)}$. We compute the reconstruction error of both directions and add them as a symmetric distance \textbf{SC}:


\vspace{-3mm}
\small
\begin{align}
 SC(S^1_q, S^2_q) & = Er(\bm{\widehat{F}_u(S^1_q)}, \bm{\widehat{F}_u(S^2_q))}  \nonumber \\
 & + Er(\bm{\widehat{F}_u(S^2_q)}, \bm{\widehat{F}_u(S^1_q)}), 
\label{eq:sc}
\end{align}
\normalsize
where $\bm{\widehat{F}_u(S_q)} = [ \frac{\underline{\vc}^q_k}{||\underline{\vc}^q_k||}  ]_{k=1...K}$ is a matrix of normalized codebook embeddings and $Er$ function is defined in~\eqref{eq:er}. We use the negative distance to represent similarity. 

\textbf{Second way}: 
One of the main challenges in unsupervised sentence similarity tasks is that we do not know 
which words are more important in each sentence. 
Intuitively, if one word in a query sentence is more important, the chance of observing related/similar words in the nearby sentences should be higher.
Thus, we should pay more attention to the words in a sentence that have higher cosine similarity with its multi-facet embeddings, a summary of the co-occurring word distribution. Specifically, our importance/attention weighting for all the words in the query sentence $S_q$ is defined by

\vspace{-3mm}
\small
\begin{align}
\bm{\underline{a}_q} &= \max(0, \bm{W(S_q)}^T \bm{\widehat{F}_u(S_q)} ) \; \underline{\textbf{1}}, 
\label{eq:attention}
\end{align}
\normalsize
where $\underline{\textbf{1}}$ is an all-one vector. We show that the attention vector (denoted as \textbf{Our a} in Table~\ref{tb:STS}) could be combined with various scoring functions and boost their performances. 
As a baseline, we also report the performance of the attention weights derived from the k-means loss rather than NNSC loss and call it \textbf{Our a (k-means)}.



\begin{figure*}[t!]
\scalebox{0.75}{
\begin{tabular}{|c|l|l|}
\hline
Sentences & \multicolumn{1}{|c|}{\colorbox{c79}{A} \colorbox{c27}{man} \colorbox{c50}{is} \colorbox{c21}{lifting} \colorbox{c23}{weights} \colorbox{c58}{in} \colorbox{c49}{a} \colorbox{c45}{garage} \colorbox{c63}{.}} & \multicolumn{1}{|c|}{\colorbox{c83}{A} \colorbox{c41}{man} \colorbox{c56}{is} \colorbox{c07}{lifting} \colorbox{c10}{weights} \colorbox{c66}{.}}   \\ \hline
& e1 \: | \space  can 0.872, even 0.851, should 0.850 & e1 \: | \space can 0.865, either 0.843, should 0.841 \\
& e2 \: | \space  front 0.762, bottom 0.742, down 0.714 & e2 \: | \space front 0.758, bottom 0.758, sides 0.691  \\
& e3 \: | \space  lifting 0.866, lift 0.663, Lifting 0.621 & e3 \: | \space lifting 0.847, lift 0.635, Lifting 0.610   \\
& e4 \: | \space  garage 0.876, garages 0.715, basement 0.707 & e4 \: | \space lifting 0.837, lift 0.652, weights 0.629   \\
Output& e5 \: | \space  decreasing 0.677, decreases 0.655, negligible 0.649 & e5 \: | \space decreasing 0.709, decreases 0.685, increases 0.682  \\
Embeddings& e6 \: | \space  weights 0.883, Weights 0.678, weight 0.665 & e6 \: | \space weights 0.864, weight 0.700, Weights 0.646  \\
& e7 \: | \space  cylindrical 0.700, plurality 0.675, axial 0.674 & e7 \: | \space annular 0.738, cylindrical 0.725, circumferential 0.701  \\
& e8 \: | \space  configurations 0.620, incorporating 0.610, utilizing 0.605 & e8 \: | \space methods 0.612, configurations 0.610, graphical 0.598  \\
& e9 \: | \space  man 0.872, woman 0.682, men 0.672 & e9 \: | \space sweating 0.498, cardiovascular 0.494, dehydration 0.485  \\
& e10 | \space  man 0.825, men 0.671, woman 0.653 & e10 | \space man 0.888, woman 0.690, men 0.676 \\ \hline
\end{tabular}
}
\centering
\caption{Comparison of our attention weights and the output embeddings between two similar sentences from STSB. A darker red indicates a larger attention value in \eqref{eq:attention} and the output embeddings are visualized using the same way in Table~\ref{tb:visual_example}.
}
\label{tb:good_topics_STS_main_paper}
\end{figure*}

\begin{table}[t!]
\scalebox{0.84}{
\begin{tabular}{|cc|cc|cc|}
\hline
\multicolumn{2}{|c|}{Method} & \multicolumn{2}{|c|}{Dev} & \multicolumn{2}{|c|}{Test} \\ \hline
\multicolumn{1}{|c|}{Score} & Model & All & Low & All & Low   \\ \hline
\multicolumn{1}{|c|}{Cosine} & Skip-thought & 43.2 & 28.1 & 30.4 & 21.2 \\ \hline
\multicolumn{1}{|c|}{CLS} & \multirow{2}{*}{BERT} & 9.6	& -0.4 & 4.1 & 0.2 \\
\multicolumn{1}{|c|}{Avg} &  &62.3 & 42.1 & 51.2 & 39.1 \\ \hline
\multirow{2}{*}{SC} & \multicolumn{1}{|c|}{Our c K1} & 55.7 & 43.7 & 47.6 & 45.4 \\
 & \multicolumn{1}{|c|}{Our c K10} & 63.0 & 51.8 & 52.6 & 47.8 \\ \hline
\multirow{3}{*}{WMD} & \multicolumn{1}{|c|}{GloVe} & 58.8 & 35.3 & 40.9 & 25.4 \\
 & \multicolumn{1}{|c|}{Our a K1} & 63.1 & 43.3 & 47.5 & 34.8 \\
 & \multicolumn{1}{|c|}{Our a K10} & 66.7 & 47.4 & 52.6 & 39.8 \\ \hline
\multirow{3}{*}{Prob\_WMD} & \multicolumn{1}{|c|}{GloVe} & 75.1 & 59.6 &	63.1 &	52.5 \\
 & \multicolumn{1}{|c|}{Our a K1} & 74.4 &	60.8 &	62.9 &	54.4 \\
 & \multicolumn{1}{|c|}{Our a K10} & \textbf{76.2} & \textbf{62.6} & \textbf{66.1} & 58.1 \\ \hline
\multirow{3}{*}{Avg} & \multicolumn{1}{|c|}{GloVe} & 51.7	& 32.8 & 36.6 & 30.9 \\
 & \multicolumn{1}{|c|}{Our a K1} & 54.5 & 40.2 & 44.1 & 40.6 \\
 & \multicolumn{1}{|c|}{Our a K10} & 61.7 & 47.1 & 50.0 & 46.5 \\ \hline
\multirow{3}{*}{Prob\_avg} & \multicolumn{1}{|c|}{GloVe} & 70.7 & 56.6 & 59.2	& 54.8 \\
& \multicolumn{1}{|c|}{Our a K1} & 68.5 & 56.0 & 58.1 & 55.2  \\
 & \multicolumn{1}{|c|}{Our a K10} & 72.0 & 60.5 & 61.4 & \textbf{59.3} \\ \hline
\multirow{4}{*}{SIF$\dagger$} & \multicolumn{1}{|c|}{GloVe} & 75.1 & 65.7 & 63.2 & 58.1 \\
 & \multicolumn{1}{|c|}{Our a K1} & 72.5 & 64.0 & 61.7 & 58.5 \\
 & \multicolumn{1}{|c|}{Our a K10} & \textbf{75.2} & \textbf{67.6} & \textbf{64.6} & \textbf{62.2} \\ 
  & \multicolumn{1}{|c|}{Our a (k-means) K10} & 71.5&	62.3&	61.5&	57.2 \\ 
 \hline \hline
\multicolumn{2}{|c|}{sentence-BERT (100 pairs)*} &71.2&55.5&64.5&58.2 \\ \hline


\end{tabular}
}
\centering
\caption{Pearson correlation (\%) in the development and test sets in the STS benchmark. The performances of all sentence pairs are indicated as All. Low means the performances on the half of sentence pairs with lower similarity (i.e., STSB Low). Our c means our codebook embeddings and Our a means our attention vectors. * indicates a supervised method. $\dagger$ indicates the methods which use training distribution to approximate testing distribution. The best score with and without $\dagger$ are highlighted.}
\label{tb:STS}
\end{table}

\textbf{Setup}: 
STS benchmark~\citep{sts} is a widely used sentence similarity task. We compare the correlations between the predicted semantic similarity and the manually labeled similarity. We report Pearson correlation coefficient, which is strongly correlated with Spearman correlation in all our experiments.
Intuitively, when two sentences are less similar to each other, humans tend to judge the similarity based on how similar their facets are. Thus, we also compare the performances on the lower half of the datasets where their ground truth similarities are less than the median similarity in the dataset, and we call this benchmark STSB Low.

A simple but effective way to measure sentence similarity is to compute the cosine similarity between the average (contextualized) word embedding~\citep{sim_sum_baseline}.
The scoring function is labeled as \textbf{Avg}. 
Besides, we test the sentence embedding from BERT and from skip-thought~\citep{kiros2015skip} (denoted as \textbf{CLS} and \textbf{Skip-thought Cosine}, respectively). 

In order to deemphasize the syntax parts of the sentences, \citet{sif} propose to weight the word $w$ in each sentence according to $\frac{\alpha}{\alpha + p(w)}$, where $\alpha$ is a constant and $p(w)$ is the probability of seeing the word $w$ in the corpus. Following its recommendation, we set $\alpha$ to be $10^{-4}$ in this paper. After the weighting, we remove the first principal component of all the sentence embeddings in the training data as suggested by \citet{sif} and denote the method as \textbf{SIF}.
The post-processing requires an estimation of testing embedding distribution, which is not desired in some applications, so we also report the performance before removing the principal component, which is called~\textbf{Prob\_avg}.  

We also test word mover's distance (\textbf{WMD}) \citep{kusner2015word}, which explicitly matches every word in a pair of sentences. As we do in \textbf{Prob\_avg}, we apply $\frac{\alpha}{\alpha + p(w)}$ to \textbf{WMD} to down-weight the importance of functional words, and call this scoring function as~\textbf{Prob\_WMD}. When using \textbf{Our a}, we multiple our attention vector with the weights of every word (e.g., $\frac{\alpha}{\alpha + p(w)}$ for \textbf{Prob\_avg} and \textbf{Prob\_WMD}).

To motivate the unsupervised setting, we present the performance of sentence-BERT \citep{reimers2019sentence} that are trained by 100 sentence pairs. We randomly sample the sentence pairs from a data source that is not included in STSB (e.g., headlines in STS 2014), and report the testing performance averaged across all the sources from STS 2012 to 2016.
More details are included in \refappsentsim.

\textbf{Results}: 
In Figure~\ref{tb:good_topics_STS_main_paper}, we first visualize our attention weights in \eqref{eq:attention} and our output codebook embeddings for a pair of similar sentences from STSB to intuitively explain why modeling co-occurring distribution could improve the similarity measurement. 

Many similar sentences might use different word choices or using extra words to describe details, but their possible nearby words are often similar. For example, appending \emph{in the garage} to \emph{A man is lifting weights} does not significantly change the facets of the sentences and thus the word \emph{garage} receives relatively a lower attention weight. This makes its similarity measurement from our methods, \textbf{Our c} and \textbf{Our a}, closer to the human judgment than other baselines.




In Table~\ref{tb:STS}, \textbf{Our c SC}, which matches between two sets of facets, outperforms \textbf{WMD}, which matches between two sets of words in the sentence, and also outperforms \textbf{BERT Avg}, especially in STSB Low. The significantly worse performances from \textbf{Skip-thought Cosine} justify our choice of ignoring the order in the co-occurring words.

All the scores in \textbf{Our * K10} are significantly better than \textbf{Our * K1}, which demonstrates the co-occurring word distribution is hard to be modeled well using a single embedding. 
Multiplying the proposed attention weighting consistently boosts the performance in all the scoring functions especially in STSB Low and without relying on the generalization assumption of the training distribution. Finally, using k-means loss, \textbf{Our a (k-means) K10}, significantly degrades the performance compared to \textbf{Our a K10}, which justify the proposed NNSC loss. In \refappsentsim, our methods are compared with more baselines using more datasets to test the effectiveness of multi-facet embeddings and our design choices more comprehensively.

\subsection{Unsupervised Extractive Summarization}
\label{sec:sum}
The classic representation of a sentence uses either a single embedding or the (contextualized) embeddings of all the words in the sentence. In this section, we would like to show that both options are not ideal for extracting a set of sentences as a document summary.

Table~\ref{tb:visual_example} indicates that our multiple codebook embeddings of a sentence capture its different facets well, so we represent a document summary $S$ as the union of the multi-facet embeddings of the sentences in the summary $R(S) = \cup_{t=1}^T \{\widehat{F}_u(S_t)\}$, where $\{\widehat{F}_u(S_t)\}$ is the set of column vectors in the matrix $\bm{\widehat{F}_u(S_t)}$ of sentence $S_t$.

A good summary should cover multiple facets that well represent all topics/concepts in the document~\citep{kobayashi2015summarization}. The objective can be quantified as discovering a summary $S$ whose multiple embeddings $R(S)$ best reconstruct the distribution of normalized word embedding $\underline{\vw}$ in the document $D$~\citep{kobayashi2015summarization}. That is,

\vspace{-3mm}
\small
\begin{align}
\argmax_{S} \sum_{\underline{\vw} \in D} \frac{\alpha}{\alpha + p(w)} \max_{\underline{\vs} \in R(S)} \underline{\vw}^T\underline{\vs},
\label{eq:sum_select}
\end{align}
\normalsize
where $\frac{\alpha}{\alpha + p(w)}$ is the weights of words we used in the sentence similarity experiments~\citep{sif}. 
We greedily select sentences to optimize~\eqref{eq:sum_select} as in~\citet{kobayashi2015summarization}.


\begin{table}[t!]
\scalebox{0.9}{
\begin{tabular}{|c|c|ccc|}
\hline
Setting & Method & R-1 & R-2 & Len \\ \hline
\multirowcell{10}{Unsup, \\ No \\ Sent \\ Order} & Random & 28.1 & 8.0 & 68.7 \\ 
& Textgraph (tfidf)$\dagger$ & 33.2  & 11.8 & -  \\ 
& Textgraph (BERT)$\dagger$ & 30.8 & 9.6 &  - \\ 
& W Emb (GloVe) & 26.6 & 8.8 & 37.0 \\ 
& Sent Emb (GloVe) & 32.6 & 10.7 & 78.2 \\ 
& W Emb (BERT) & 31.3 & 11.2 & 45.0 \\ 
& Sent Emb (BERT) & 32.3  & 10.6 & 91.2 \\ 
& Our c (K=3) & 32.2 & 10.1 & 75.4 \\ 
& Our c (K=10) & 34.0 & 11.6 & 81.3 \\ 
& Our c (K=100) & \textbf{35.0} & \textbf{12.8} & 92.9 \\ \hline
\multirow{2}{*}{Unsup} & Lead-3 & 40.3 & 17.6 & 87.0  \\ 
& PACSUM (BERT)$\dagger$ & \textbf{40.7}  & \textbf{17.8}  & -  \\ \hline
Sup & RL* &  \textbf{41.7} & \textbf{19.5}  & - \\ \hline
\end{tabular}
}
\centering
\caption{The ROUGE F1 scores of different methods on CNN/Daily Mail dataset. The results with $\dagger$ are taken from~\citet{zheng2019sentence}. The results with * are taken from~\citet{celikyilmaz2018deep}. }
\label{tb:sum}
\end{table}

\textbf{Setup}:
We compare our multi-facet embeddings with other alternative ways of modeling the facets of sentences. A simple way is to compute the average word embedding as a single-facet sentence embedding.\footnote{Although \eqref{eq:sum_select} weights each word in the document, we find that the weighting $\frac{\alpha}{\alpha + p(w)}$ does not improve the sentence representation when averaging the word embeddings.}
This baseline is labeled as \textbf{Sent Emb}. Another way is to use the (contextualized) embedding of all the words in the sentences as different facets of the sentences. Since longer sentences have more words, we normalize the gain of the reconstruction similarity by the sentence length. The method is denoted as \textbf{W Emb}.
We also test the baselines of selecting random sentences (\textbf{Rnd}) and first 3 sentences (\textbf{Lead-3}) in the document.

The results on the testing set of CNN/Daily Mail~\citep{cnn_dataset,cnn_dataset_split} are compared using F1 of ROUGE~\citep{rouge} in Table~\ref{tb:sum}. R-1, R-2, and Len mean ROUGE-1, ROUGE-2, and average summary length, respectively. All methods choose 3 sentences by following the setting in \citet{zheng2019sentence}. \emph{Unsup, No Sent Order} means the methods do not use the sentence order information in CNN/Daily Mail.

In CNN/Daily Mail, the unsupervised methods which access sentence order information such as \textbf{Lead-3} have performances similar to supervised methods such as RL~\citep{celikyilmaz2018deep}. To evaluate the quality of unsupervised sentence embeddings, we focus on comparing the unsupervised methods which do not assume the first few sentences form a good summary.



\textbf{Results}:
In Table~\ref{tb:sum}, predicting 100 clusters yields the best results. Notice that our method greatly alleviates the computational and sample efficiency challenges, which allows us to set cluster numbers $K$ to be a relatively large number.

The results highlight the limitation of classic representations. The single sentence embedding cannot capture its multiple facets. On the other hand, if a sentence is represented by the embeddings of its words, it is difficult to eliminate the bias of selecting longer or shorter sentences and a facet might be composed by multiple words (e.g., the input sentence in Table~\ref{tb:visual_example} describes a service, but there is not a single word in the sentence that means service).

\subsection{Unsupervised Phrase Similarity}
Recently, \citet{dubossarsky2018coming} discovered that the multiple embeddings of each word may not improve the performance in word similarity benchmarks even if they capture more senses or facets of polysemies.
Since our method can improve the sentence similarity estimation, we want to see whether multi-facet embeddings could also help the phrase similarity estimation.

In addition to \textbf{SC} in \eqref{eq:sc}, we also compute the average of the contextualized word embeddings from our transformer encoder as the phrase embedding. We find that the cosine similarity between the two phrase embeddings is a good similarity estimation, and the method is labeled as \textbf{Ours Emb}. 

\begin{table}[t!]
\scalebox{0.87}{
\begin{tabular}{|cc|cc|c|c|}
\hline
\multicolumn{2}{|c|}{Method} & \multicolumn{2}{c|}{SemEval 2013} & Turney (5) & Turney (10)   \\ \hline 
\multicolumn{1}{|c|}{Model} & Score & AUC & F1 & Accuracy & Accuracy \\ \hline
\multirow{2}{*}{BERT} & \multicolumn{1}{|c|}{CLS} & 54.7 & 66.7 & 29.2 & 15.5  \\ 
 & \multicolumn{1}{|c|}{Avg} & 66.5 & 67.1 & 43.4 & 24.3  \\ \hline
GloVe & \multicolumn{1}{|c|}{Avg} & 79.5 & 73.7 & 25.9 & 12.9 \\ \hline
\multicolumn{1}{|c|}{FCT LM$\dagger$} & Emb & - & 67.2 & 42.6 & 27.6 \\ \hline
Ours& \multicolumn{1}{|c|}{SC} & 80.3 & 72.8 & 45.6 & 28.8   \\ 
(K=10) & \multicolumn{1}{|c|}{Emb} & 85.6 & 77.1 & 49.4 & 31.8  \\ \hline
Ours & \multicolumn{1}{|c|}{SC} & 81.1 & 72.7 & 45.3 & 28.4   \\ 
(K=1) & \multicolumn{1}{|c|}{Emb} & \textbf{87.8} & \textbf{78.6} & \textbf{50.3} & \textbf{32.5}  \\ \hline
\end{tabular}
}
\centering
\caption{Performance of phrase similarity tasks. Every model is trained on a lowercased corpus. In SemEval 2013, AUC (\%) is the area under the precision-recall curve of classifying similar phrase pairs. In Turney, we report the accuracy (\%) of predicting the correct similar phrase pair among 5 or 10 candidate pairs. The results with $\dagger$ are taken from~\citet{fct}. }
\label{tb:phrase_sim}
\end{table}

\textbf{Setup}:
We evaluate our phrase similarity using SemEval 2013 task 5(a) English~\citep{semeval2013} and Turney 2012~\citep{turney2012}. The task of SemEval 2013 is to distinguish similar phrase pairs from dissimilar phrase pairs. In Turney (5), given each query bigram, each model predicts the most similar unigram among 5 candidates, and Turney (10) adds 5 more negative phrase pairs by pairing the reverse of the query bigram with the 5 unigrams. 


\textbf{Results}:
The performances are presented in Table~\ref{tb:phrase_sim}. \textbf{Ours (K=1)} is usually slightly better than \textbf{Ours (K=10)}, and the result supports the finding of~\citet{dubossarsky2018coming}. We hypothesize that unlike sentences, most of the phrases have only one facet/sense, and thus can be modeled by a single embedding well. In \refapphyper, the hypernym detection results also support this hypothesis.

Even though being slightly worse, the performances of \textbf{Ours (K=10)} remain strong compared with baselines. This implies that the similarity performances are not sensitive to the number of clusters as long as sufficiently large K is used because the model is able to output multiple nearly duplicated codebook embeddings to represent one facet (e.g., using two centers to represent the facet related to \emph{company} in Figure~\ref{fig:first_page}). The flexibility alleviates the issues of selecting K in practice. Finally, the strong performances in Turney (10) verify that our encoder respects the word order when composing the input sequence.

\section{Related Work}
\label{sec:related}

Topic modeling~\citep{blei2003latent} has been extensively studied and widely applied due to its interpretability and flexibility of incorporating different forms of input features~\citep{MimnoM08}. 
\citet{cao2015novel, SrivastavaS17} demonstrate that neural networks could be applied to discover semantically coherent topics. Instead of optimizing a global topic model, our goal is to efficiently discover different sets of topics/clusters on the words beside each (unseen) phrase or sentence.

Recently, \citet{gupta2019improving} and \citet{gupta2020p} discover that global clustering could improve the representation of sentences and documents. In our work, we show that a local clustering could be used in several downstream applications, including word importance estimation for measuring sentence similarity. Whether combining global clustering and local clustering could lead to a further improvement is an interesting future research direction.

Sparse coding on word embedding space is used to model the multiple facets of a word~\citep{faruqui2015sparse,arora2018linear}, and parameterizing word embeddings using neural networks is used to test hypothesis~\citep{HanGSC18} and save storage space~\citep{ShuN18}. Besides, to capture asymmetric relations such as hypernyms, words are represented as single or multiple regions in Gaussian embeddings~\citep{VilnisM15, athiwaratkun2017multimodal} rather than a single point. However, the challenges of extending these methods to longer sequences are not addressed in these studies.

One of our main challenges is to design a loss for learning to predict cluster centers while modeling the dependency among the clusters. This requires a matching step between two sets and computing the distance loss after the matching~\citep{eiter1997distance}. One popular loss is called Chamfer distance, which is widely adopted in the auto-encoder models for point clouds~\citep{yang2018foldingnet, liu2019l2g}, while more sophisticated matching loss options are also proposed~\citep{stewart2016end, balles2019holographic}. The goal of the previous studies focuses on measuring symmetric distances between the ground truth set and predicted set (usually with an equal size), while our loss tries to reconstruct the ground truth set using much fewer codebook embeddings. 

Other ways to achieve the permutation invariant loss for neural networks include sequential decision making~\citep{WelleckYGMZC18}, mixture of experts~\citep{yang2018breaking, wang2019attention}, beam search~\citep{label_set}, predicting the permutation using a CNN~\citep{rezatofighi2018deep}, Transformers~\citep{stern2019insertion, gu2019insertion, carion2020} or reinforcement learning~\citep{welleck2019non}. In contrast, our goal is to efficiently predict a set of cluster centers that can well reconstruct the set of observed instances rather than directly predicting the observed instances.












\section{Conclusions}
In this work, we propose a framework for learning the co-occurring distribution of the words surrounding a sentence or a phrase.
Even though there are usually only a few words that co-occur with each sentence, we demonstrate that the proposed models can learn to predict interpretable cluster centers conditioned on an (unseen) sentence.

In the sentence similarity tasks, the results indicate that the similarity between two sets of multi-facet embeddings well correlates with human judgments, and we can use the multi-facet embeddings to estimate the word importance and improve various widely-used similarity measurements in a pre-trained word embedding space such as GloVe. In a single-document extractive summarization task, we demonstrate multi-facet embeddings significantly outperform classic unsupervised sentence embedding or individual word embeddings. 
Finally, the results of phrase similarity tasks suggest that a single embedding might be sufficient to represent the co-occurring word distribution of a phrase.

\section*{Acknowledgements}
We thank Ao Liu and Mohit Iyyer for many helpful discussions and Nishant Yadav for suggesting several related work. We also thank the anonymous reviewers for their constructive feedback.

This work was supported in part by the Center for Data Science and the Center for Intelligent Information Retrieval, in part by the Chan Zuckerberg Initiative under the project Scientific Knowledge Base Construction, in part using high performance computing equipment obtained under a grant from the Collaborative R\&D Fund managed by the Massachusetts Technology Collaborative, in part by the National Science Foundation (NSF) grant numbers DMR-1534431 and IIS-1514053. 

Any opinions, findings, conclusions, or recommendations expressed in this material are those of the authors and do not necessarily reflect those of the sponsor.

\section*{Ethics Statement}
We propose a novel framework, neural architecture, and loss to learn multi-facet embedding from the co-occurring statistics in NLP. In this study, we exploit the co-occurring relation between a sentence and its nearby words to improve the sentence representation. In our follow-up studies, we discover that the multi-facet embeddings could also be used to learn other types of co-occurring statistics. For example, we can use the co-occurring relation between a scientific paper and its citing paper to improve paper recommendation methods in \citet{BansalBM16}. \citet{our_eacl_re} use the co-occurring relation between a sentence pattern and its entity pair to improve relation extraction in \citet{verga2016multilingual}. \citet{our_eacl_interactive} use the co-occurring relation between a context paragraph and its subsequent words to control the topics of language generation. In the future, the approach might also be used to improve the efficiency of document similarity estimation~\cite{luan2020sparse}.



On the other hand, we improve the sentence similarity and summarization tasks in this work using the assumption that important words are more likely to appear in the nearby sentences. The assumption might be violated in some domains and our method might degrade the performances in such domains if the practitioner applies our methods without considering the validity of the assumption.

\bibliography{ref}

\begin{thebibliography}{76}
\providecommand{\natexlab}[1]{#1}
\providecommand{\url}[1]{\texttt{#1}}
\providecommand{\urlprefix}{URL }
\expandafter\ifx\csname urlstyle\endcsname\relax
  \providecommand{\doi}[1]{doi:\discretionary{}{}{}#1}\else
  \providecommand{\doi}{doi:\discretionary{}{}{}\begingroup
  \urlstyle{rm}\Url}\fi

\bibitem[{Agirre et~al.(2015)Agirre, Banea, Cardie, Cer, Diab, Gonzalez-Agirre,
  Guo, Lopez-Gazpio, Maritxalar, Mihalcea, Rigau, Uria, and
  Wiebe}]{agirre2015semeval}
Agirre, E.; Banea, C.; Cardie, C.; Cer, D.; Diab, M.; Gonzalez-Agirre, A.; Guo,
  W.; Lopez-Gazpio, I.; Maritxalar, M.; Mihalcea, R.; Rigau, G.; Uria, L.; and
  Wiebe, J. 2015.
\newblock Semeval-2015 task 2: Semantic textual similarity, english, spanish
  and pilot on interpretability.
\newblock In \emph{SemEval}.

\bibitem[{Agirre et~al.(2014)Agirre, Banea, Cardie, Cer, Diab, Gonzalez-Agirre,
  Guo, Mihalcea, Rigau, and Wiebe}]{agirre2014semeval}
Agirre, E.; Banea, C.; Cardie, C.; Cer, D.; Diab, M.; Gonzalez-Agirre, A.; Guo,
  W.; Mihalcea, R.; Rigau, G.; and Wiebe, J. 2014.
\newblock Semeval-2014 task 10: Multilingual semantic textual similarity.
\newblock In \emph{SemEval}.

\bibitem[{Agirre et~al.(2016)Agirre, Banea, Cer, Diab, Gonzalez-Agirre,
  Mihalcea, Rigau, and Wiebe}]{agirre2016semeval}
Agirre, E.; Banea, C.; Cer, D.; Diab, M.; Gonzalez-Agirre, A.; Mihalcea, R.;
  Rigau, G.; and Wiebe, J. 2016.
\newblock Semeval-2016 task 1: Semantic textual similarity, monolingual and
  cross-lingual evaluation.
\newblock In \emph{SemEval}.

\bibitem[{Agirre et~al.(2013)Agirre, Cer, Diab, Gonzalez-Agirre, and
  Guo}]{agirre2013sem}
Agirre, E.; Cer, D.; Diab, M.; Gonzalez-Agirre, A.; and Guo, W. 2013.
\newblock * SEM 2013 shared task: Semantic textual similarity.
\newblock In \emph{* SEM}.

\bibitem[{Agirre et~al.(2012)Agirre, Diab, Cer, and
  Gonzalez-Agirre}]{agirre2012semeval}
Agirre, E.; Diab, M.; Cer, D.; and Gonzalez-Agirre, A. 2012.
\newblock Semeval-2012 task 6: A pilot on semantic textual similarity.
\newblock In \emph{SemEval}.

\bibitem[{Arora et~al.(2018)Arora, Li, Liang, Ma, and
  Risteski}]{arora2018linear}
Arora, S.; Li, Y.; Liang, Y.; Ma, T.; and Risteski, A. 2018.
\newblock Linear algebraic structure of word senses, with applications to
  polysemy.
\newblock \emph{Transactions of the Association of Computational Linguistics}
  6: 483--495.

\bibitem[{Arora, Liang, and Ma(2017)}]{sif}
Arora, S.; Liang, Y.; and Ma, T. 2017.
\newblock A Simple but Tough-to-beat Baseline for Sentence Embeddings.
\newblock In \emph{ICLR}.

\bibitem[{Asaadi, Mohammad, and Kiritchenko(2019)}]{bird-naacl2019}
Asaadi, S.; Mohammad, S.~M.; and Kiritchenko, S. 2019.
\newblock Big BiRD: A Large, Fine-Grained, Bigram Relatedness Dataset for
  Examining Semantic Composition.
\newblock In \emph{NAACL-HLT}.

\bibitem[{Athiwaratkun and Wilson(2017)}]{athiwaratkun2017multimodal}
Athiwaratkun, B.; and Wilson, A. 2017.
\newblock Multimodal Word Distributions.
\newblock In \emph{ACL}.

\bibitem[{Balles and Fischbacher(2019)}]{balles2019holographic}
Balles, L.; and Fischbacher, T. 2019.
\newblock Holographic and other Point Set Distances for Machine Learning.
\newblock \urlprefix\url{https://openreview.net/forum?id=rJlpUiAcYX}.

\bibitem[{Bansal, Belanger, and McCallum(2016)}]{BansalBM16}
Bansal, T.; Belanger, D.; and McCallum, A. 2016.
\newblock Ask the {GRU}: Multi-task Learning for Deep Text Recommendations.
\newblock In \emph{RecSys}.

\bibitem[{Bentley(1975)}]{bentley1975multidimensional}
Bentley, J.~L. 1975.
\newblock Multidimensional binary search trees used for associative searching.
\newblock \emph{Communications of the ACM} 18(9): 509--517.

\bibitem[{Bird, Klein, and Loper(2009)}]{bird2009natural}
Bird, S.; Klein, E.; and Loper, E. 2009.
\newblock \emph{Natural language processing with Python: analyzing text with
  the natural language toolkit}.
\newblock " O'Reilly Media, Inc.".

\bibitem[{Blei, Ng, and Jordan(2003)}]{blei2003latent}
Blei, D.~M.; Ng, A.~Y.; and Jordan, M.~I. 2003.
\newblock Latent dirichlet allocation.
\newblock \emph{Journal of machine Learning research} 3(Jan): 993--1022.

\bibitem[{Bojanowski et~al.(2017)Bojanowski, Grave, Joulin, and
  Mikolov}]{bojanowski2017enriching}
Bojanowski, P.; Grave, E.; Joulin, A.; and Mikolov, T. 2017.
\newblock Enriching word vectors with subword information.
\newblock \emph{Transactions of the Association for Computational Linguistics}
  5: 135--146.

\bibitem[{Cao et~al.(2015)Cao, Li, Liu, Li, and Ji}]{cao2015novel}
Cao, Z.; Li, S.; Liu, Y.; Li, W.; and Ji, H. 2015.
\newblock A novel neural topic model and its supervised extension.
\newblock In \emph{AAAI}.

\bibitem[{Carion et~al.(2020)Carion, Massa, Synnaeve, Usunier, Kirillov, and
  Zagoruyko}]{carion2020}
Carion, N.; Massa, F.; Synnaeve, G.; Usunier, N.; Kirillov, A.; and Zagoruyko,
  S. 2020.
\newblock End-to-End Object Detection with Transformers.
\newblock \emph{arXiv preprint arXiv:2005.12872} .

\bibitem[{Celikyilmaz et~al.(2018)Celikyilmaz, Bosselut, He, and
  Choi}]{celikyilmaz2018deep}
Celikyilmaz, A.; Bosselut, A.; He, X.; and Choi, Y. 2018.
\newblock Deep Communicating Agents for Abstractive Summarization.
\newblock In \emph{NAACL-HLT}.

\bibitem[{Cer et~al.(2017)Cer, Diab, Agirre, Lopez-Gazpio, and Specia}]{sts}
Cer, D.; Diab, M.; Agirre, E.; Lopez-Gazpio, I.; and Specia, L. 2017.
\newblock SemEval-2017 Task 1: Semantic Textual Similarity Multilingual and
  Crosslingual Focused Evaluation.
\newblock In \emph{SemEval-2017}.

\bibitem[{Chang et~al.(2021)Chang, Yuan, Iyyer, and
  McCallum}]{our_eacl_interactive}
Chang, H.-S.; Yuan, J.; Iyyer, M.; and McCallum, A. 2021.
\newblock Changing the Mind of Transformers for Topically-Controllable Language
  Generation.
\newblock In \emph{EACL}.

\bibitem[{Devlin et~al.(2019)Devlin, Chang, Lee, and Toutanova}]{BERT}
Devlin, J.; Chang, M.; Lee, K.; and Toutanova, K. 2019.
\newblock {BERT:} Pre-training of Deep Bidirectional Transformers for Language
  Understanding.
\newblock In \emph{NAACL-HLT}.

\bibitem[{Dubossarsky, Grossman, and Weinshall(2018)}]{dubossarsky2018coming}
Dubossarsky, H.; Grossman, E.; and Weinshall, D. 2018.
\newblock Coming to your senses: on controls and evaluation sets in polysemy
  research.
\newblock In \emph{EMNLP}.

\bibitem[{Eiter and Mannila(1997)}]{eiter1997distance}
Eiter, T.; and Mannila, H. 1997.
\newblock Distance measures for point sets and their computation.
\newblock \emph{Acta Informatica} 34(2): 109--133.

\bibitem[{Faruqui et~al.(2015)Faruqui, Tsvetkov, Yogatama, Dyer, and
  Smith}]{faruqui2015sparse}
Faruqui, M.; Tsvetkov, Y.; Yogatama, D.; Dyer, C.; and Smith, N.~A. 2015.
\newblock Sparse Overcomplete Word Vector Representations.
\newblock In \emph{ACL}.

\bibitem[{Gu, Liu, and Cho(2019)}]{gu2019insertion}
Gu, J.; Liu, Q.; and Cho, K. 2019.
\newblock Insertion-based decoding with automatically inferred generation
  order.
\newblock \emph{Transactions of the Association for Computational Linguistics}
  7: 661--676.

\bibitem[{Gupta et~al.(2019)Gupta, Saw, Nokhiz, Gupta, and
  Talukdar}]{gupta2019improving}
Gupta, V.; Saw, A.; Nokhiz, P.; Gupta, H.; and Talukdar, P. 2019.
\newblock Improving document classification with multi-sense embeddings.
\newblock In \emph{ECAI}.

\bibitem[{Gupta et~al.(2020)Gupta, Saw, Nokhiz, Netrapalli, Rai, and
  Talukdar}]{gupta2020p}
Gupta, V.; Saw, A.; Nokhiz, P.; Netrapalli, P.; Rai, P.; and Talukdar, P. 2020.
\newblock {P-SIF}: Document embeddings using partition averaging.
\newblock In \emph{AAAI}.

\bibitem[{Han et~al.(2018)Han, Gill, Spirling, and Cho}]{HanGSC18}
Han, R.; Gill, M.; Spirling, A.; and Cho, K. 2018.
\newblock Conditional Word Embedding and Hypothesis Testing via
  Bayes-by-Backprop.
\newblock In \emph{EMNLP}.

\bibitem[{Hermann et~al.(2015)Hermann, Kocisky, Grefenstette, Espeholt, Kay,
  Suleyman, and Blunsom}]{cnn_dataset}
Hermann, K.~M.; Kocisky, T.; Grefenstette, E.; Espeholt, L.; Kay, W.; Suleyman,
  M.; and Blunsom, P. 2015.
\newblock Teaching machines to read and comprehend.
\newblock In \emph{NeurIPS}.

\bibitem[{Hoyer(2002)}]{hoyer2002non}
Hoyer, P.~O. 2002.
\newblock Non-negative Sparse Coding.
\newblock In \emph{Proceedings of the 12th IEEE Workshop on Neural Networks for
  Signal Processing}.

\bibitem[{Huang, Ji et~al.(2017)}]{huang2017learning}
Huang, L.; Ji, H.; et~al. 2017.
\newblock Learning Phrase Embeddings from Paraphrases with GRUs.
\newblock In \emph{Proceedings of the First Workshop on Curation and
  Applications of Parallel and Comparable Corpora}.

\bibitem[{Kiros et~al.(2015)Kiros, Zhu, Salakhutdinov, Zemel, Urtasun,
  Torralba, and Fidler}]{kiros2015skip}
Kiros, R.; Zhu, Y.; Salakhutdinov, R.~R.; Zemel, R.; Urtasun, R.; Torralba, A.;
  and Fidler, S. 2015.
\newblock Skip-thought vectors.
\newblock In \emph{NeurIPS}.

\bibitem[{Kobayashi, Noguchi, and Yatsuka(2015)}]{kobayashi2015summarization}
Kobayashi, H.; Noguchi, M.; and Yatsuka, T. 2015.
\newblock Summarization based on embedding distributions.
\newblock In \emph{EMNLP}.

\bibitem[{Korkontzelos et~al.(2013)Korkontzelos, Zesch, Zanzotto, and
  Biemann}]{semeval2013}
Korkontzelos, I.; Zesch, T.; Zanzotto, F.~M.; and Biemann, C. 2013.
\newblock Semeval-2013 task 5: Evaluating phrasal semantics.
\newblock In \emph{SemEval 2013}.

\bibitem[{Kumar and Tsvetkov(2019)}]{kumar2018mises}
Kumar, S.; and Tsvetkov, Y. 2019.
\newblock {Von Mises-Fisher} Loss for Training Sequence to Sequence Models with
  Continuous Outputs.
\newblock In \emph{ICLR}.

\bibitem[{Kusner et~al.(2015)Kusner, Sun, Kolkin, and
  Weinberger}]{kusner2015word}
Kusner, M.; Sun, Y.; Kolkin, N.; and Weinberger, K. 2015.
\newblock From word embeddings to document distances.
\newblock In \emph{ICML}.

\bibitem[{Lau et~al.(2012)Lau, Cook, McCarthy, Newman, and
  Baldwin}]{lau2012word}
Lau, J.~H.; Cook, P.; McCarthy, D.; Newman, D.; and Baldwin, T. 2012.
\newblock Word sense induction for novel sense detection.
\newblock In \emph{EACL}.

\bibitem[{Lee et~al.(2019)Lee, Lee, Kim, Kosiorek, Choi, and Teh}]{lee2018set}
Lee, J.; Lee, Y.; Kim, J.; Kosiorek, A.~R.; Choi, S.; and Teh, Y.~W. 2019.
\newblock Set transformer: A framework for attention-based
  permutation-invariant neural networks.
\newblock In \emph{ICML}.

\bibitem[{Li et~al.(2019)Li, Chen, Hsieh, and Chang}]{li2019efficient}
Li, L.~H.; Chen, P.~H.; Hsieh, C.-J.; and Chang, K.-W. 2019.
\newblock Efficient Contextual Representation Learning With Continuous Outputs.
\newblock \emph{Transactions of the Association for Computational Linguistics}
  7: 611--624.

\bibitem[{Lin and Hovy(2003)}]{rouge}
Lin, C.-Y.; and Hovy, E. 2003.
\newblock Automatic evaluation of summaries using n-gram co-occurrence
  statistics.
\newblock In \emph{NAACL-HLT}.

\bibitem[{Liu et~al.(2019)Liu, Han, Wen, Liu, and Zwicker}]{liu2019l2g}
Liu, X.; Han, Z.; Wen, X.; Liu, Y.-S.; and Zwicker, M. 2019.
\newblock L2g auto-encoder: Understanding point clouds by local-to-global
  reconstruction with hierarchical self-attention.
\newblock In \emph{Proceedings of the 27th ACM International Conference on
  Multimedia}.

\bibitem[{Luan et~al.(2020)Luan, Eisenstein, Toutanova, and
  Collins}]{luan2020sparse}
Luan, Y.; Eisenstein, J.; Toutanova, K.; and Collins, M. 2020.
\newblock Sparse, Dense, and Attentional Representations for Text Retrieval.
\newblock \emph{arXiv preprint arXiv:2005.00181} .

\bibitem[{Mikolov et~al.(2013)Mikolov, Sutskever, Chen, Corrado, and
  Dean}]{word2vec}
Mikolov, T.; Sutskever, I.; Chen, K.; Corrado, G.; and Dean, J. 2013.
\newblock Distributed representations of words and phrases and their
  compositionality.
\newblock In \emph{NeurIPS}.

\bibitem[{Milajevs et~al.(2014)Milajevs, Kartsaklis, Sadrzadeh, and
  Purver}]{sim_sum_baseline}
Milajevs, D.; Kartsaklis, D.; Sadrzadeh, M.; and Purver, M. 2014.
\newblock Evaluating Neural Word Representations in Tensor-Based Compositional
  Settings.
\newblock In \emph{EMNLP}.

\bibitem[{Mimno and McCallum(2008)}]{MimnoM08}
Mimno, D.~M.; and McCallum, A. 2008.
\newblock Topic Models Conditioned on Arbitrary Features with
  Dirichlet-multinomial Regression.
\newblock In \emph{UAI}.

\bibitem[{Neelakantan et~al.(2014)Neelakantan, Shankar, Passos, and
  McCallum}]{NeelakantanSPM14}
Neelakantan, A.; Shankar, J.; Passos, A.; and McCallum, A. 2014.
\newblock Efficient Non-parametric Estimation of Multiple Embeddings per Word
  in Vector Space.
\newblock In \emph{EMNLP}.

\bibitem[{Newman-Griffis, Lai, and Fosler-Lussier(2018)}]{WikiSRS}
Newman-Griffis, D.; Lai, A.~M.; and Fosler-Lussier, E. 2018.
\newblock Jointly Embedding Entities and Text with Distant Supervision.
\newblock In \emph{Proceedings of the 3rd Workshop on Representation Learning
  for NLP (Repl4NLP)}.

\bibitem[{Pagliardini, Gupta, and
  Jaggi(2018{\natexlab{a}})}]{pagliardini2018unsupervised}
Pagliardini, M.; Gupta, P.; and Jaggi, M. 2018{\natexlab{a}}.
\newblock Unsupervised Learning of Sentence Embeddings using Compositional
  n-Gram Features.
\newblock In \emph{NAACL-HLT}.

\bibitem[{Pagliardini, Gupta, and Jaggi(2018{\natexlab{b}})}]{pgj2017unsup}
Pagliardini, M.; Gupta, P.; and Jaggi, M. 2018{\natexlab{b}}.
\newblock Unsupervised Learning of Sentence Embeddings using Compositional
  n-Gram Features.
\newblock In \emph{NAACL}.

\bibitem[{Paul, Chang, and McCallum(2021)}]{our_eacl_re}
Paul, R.; Chang, H.-S.; and McCallum, A. 2021.
\newblock Multi-facet Universal Schema.
\newblock In \emph{EACL}.

\bibitem[{Pavlick et~al.(2015)Pavlick, Rastogi, Ganitkevitch, Van~Durme, and
  Callison-Burch}]{pavlick2015ppdb}
Pavlick, E.; Rastogi, P.; Ganitkevitch, J.; Van~Durme, B.; and Callison-Burch,
  C. 2015.
\newblock PPDB 2.0: Better paraphrase ranking, fine-grained entailment
  relations, word embeddings, and style classification.
\newblock In \emph{ACL}.

\bibitem[{Pennington, Socher, and Manning(2014)}]{glove}
Pennington, J.; Socher, R.; and Manning, C. 2014.
\newblock {GloVe}: Global vectors for word representation.
\newblock In \emph{EMNLP}.

\bibitem[{Peters et~al.(2018)Peters, Neumann, Iyyer, Gardner, Clark, Lee, and
  Zettlemoyer}]{ELMo}
Peters, M.~E.; Neumann, M.; Iyyer, M.; Gardner, M.; Clark, C.; Lee, K.; and
  Zettlemoyer, L. 2018.
\newblock Deep contextualized word representations.
\newblock In \emph{NAACL-HLT}.

\bibitem[{Qin et~al.(2019)Qin, Li, Pavlu, and Aslam}]{label_set}
Qin, K.; Li, C.; Pavlu, V.; and Aslam, J.~A. 2019.
\newblock Adapting RNN Sequence Prediction Model to Multi-label Set Prediction.
\newblock In \emph{NAACL-HLT}.

\bibitem[{Reimers and Gurevych(2019)}]{reimers2019sentence}
Reimers, N.; and Gurevych, I. 2019.
\newblock Sentence-BERT: Sentence Embeddings using Siamese BERT-Networks.
\newblock In \emph{EMNLP-IJCNLP}.

\bibitem[{Rezatofighi et~al.(2018)Rezatofighi, Kaskman, Motlagh, Shi, Cremers,
  Leal-Taix{\'e}, and Reid}]{rezatofighi2018deep}
Rezatofighi, S.~H.; Kaskman, R.; Motlagh, F.~T.; Shi, Q.; Cremers, D.;
  Leal-Taix{\'e}, L.; and Reid, I. 2018.
\newblock Deep perm-set net: learn to predict sets with unknown permutation and
  cardinality using deep neural networks.
\newblock \emph{arXiv preprint arXiv:1805.00613} .

\bibitem[{See, Liu, and Manning(2017)}]{cnn_dataset_split}
See, A.; Liu, P.~J.; and Manning, C.~D. 2017.
\newblock Get To The Point: Summarization with Pointer-Generator Networks.
\newblock In \emph{ACL}.

\bibitem[{Shu and Nakayama(2018)}]{ShuN18}
Shu, R.; and Nakayama, H. 2018.
\newblock Compressing Word Embeddings via Deep Compositional Code Learning.
\newblock In \emph{ICLR}.

\bibitem[{Shwartz, Goldberg, and Dagan(2016)}]{hypenet}
Shwartz, V.; Goldberg, Y.; and Dagan, I. 2016.
\newblock Improving Hypernymy Detection with an Integrated Path-based and
  Distributional Method.
\newblock In \emph{ACL}.

\bibitem[{Singh et~al.(2020)Singh, Hug, Dieuleveut, and
  Jaggi}]{singh2020context}
Singh, S.~P.; Hug, A.; Dieuleveut, A.; and Jaggi, M. 2020.
\newblock Context mover’s distance \& barycenters: Optimal transport of
  contexts for building representations.
\newblock In \emph{International Conference on Artificial Intelligence and
  Statistics}.

\bibitem[{Srivastava and Sutton(2017)}]{SrivastavaS17}
Srivastava, A.; and Sutton, C.~A. 2017.
\newblock Autoencoding Variational Inference For Topic Models.
\newblock In \emph{ICLR}.

\bibitem[{Stern et~al.(2019)Stern, Chan, Kiros, and
  Uszkoreit}]{stern2019insertion}
Stern, M.; Chan, W.; Kiros, J.; and Uszkoreit, J. 2019.
\newblock Insertion Transformer: Flexible Sequence Generation via Insertion
  Operations.
\newblock In \emph{ICML}.

\bibitem[{Stewart, Andriluka, and Ng(2016)}]{stewart2016end}
Stewart, R.; Andriluka, M.; and Ng, A.~Y. 2016.
\newblock End-to-end people detection in crowded scenes.
\newblock In \emph{CVPR}.

\bibitem[{Sutskever, Vinyals, and Le(2014)}]{sutskever2014sequence}
Sutskever, I.; Vinyals, O.; and Le, Q.~V. 2014.
\newblock Sequence to sequence learning with neural networks.
\newblock In \emph{NeurIPS}.

\bibitem[{Tieleman and Hinton(2012)}]{tieleman2012lecture}
Tieleman, T.; and Hinton, G. 2012.
\newblock Lecture 6.5-rmsprop: Divide the gradient by a running average of its
  recent magnitude.
\newblock \emph{COURSERA: Neural networks for machine learning} 4(2): 26--31.

\bibitem[{Turney(2012)}]{turney2012}
Turney, P.~D. 2012.
\newblock Domain and function: A dual-space model of semantic relations and
  compositions.
\newblock \emph{Journal of Artificial Intelligence Research} .

\bibitem[{Vaswani et~al.(2017)Vaswani, Shazeer, Parmar, Uszkoreit, Jones,
  Gomez, Kaiser, and Polosukhin}]{vaswani2017attention}
Vaswani, A.; Shazeer, N.; Parmar, N.; Uszkoreit, J.; Jones, L.; Gomez, A.~N.;
  Kaiser, {\L}.; and Polosukhin, I. 2017.
\newblock Attention is all you need.
\newblock In \emph{NeurIPS}.

\bibitem[{Verga et~al.(2016)Verga, Belanger, Strubell, Roth, and
  McCallum}]{verga2016multilingual}
Verga, P.; Belanger, D.; Strubell, E.; Roth, B.; and McCallum, A. 2016.
\newblock Multilingual Relation Extraction using Compositional Universal
  Schema.
\newblock In \emph{NAACL-HLT}.

\bibitem[{Vilnis and McCallum(2015)}]{VilnisM15}
Vilnis, L.; and McCallum, A. 2015.
\newblock Word Representations via Gaussian Embedding.
\newblock In \emph{ICLR}.

\bibitem[{Wang, Cho, and Wen(2019)}]{wang2019attention}
Wang, T.; Cho, K.; and Wen, M. 2019.
\newblock Attention-based mixture density recurrent networks for history-based
  recommendation.
\newblock In \emph{Proceedings of the 1st International Workshop on Deep
  Learning Practice for High-Dimensional Sparse Data}.

\bibitem[{Welleck et~al.(2019)Welleck, Brantley, Daum{\'e}~III, and
  Cho}]{welleck2019non}
Welleck, S.; Brantley, K.; Daum{\'e}~III, H.; and Cho, K. 2019.
\newblock Non-Monotonic Sequential Text Generation.
\newblock In \emph{ICML}.

\bibitem[{Welleck et~al.(2018)Welleck, Yao, Gai, Mao, Zhang, and
  Cho}]{WelleckYGMZC18}
Welleck, S.; Yao, Z.; Gai, Y.; Mao, J.; Zhang, Z.; and Cho, K. 2018.
\newblock Loss Functions for Multiset Prediction.
\newblock In \emph{NeurIPS}.

\bibitem[{Yang et~al.(2018{\natexlab{a}})Yang, Feng, Shen, and
  Tian}]{yang2018foldingnet}
Yang, Y.; Feng, C.; Shen, Y.; and Tian, D. 2018{\natexlab{a}}.
\newblock Foldingnet: Point cloud auto-encoder via deep grid deformation.
\newblock In \emph{CVPR}.

\bibitem[{Yang et~al.(2018{\natexlab{b}})Yang, Dai, Salakhutdinov, and
  Cohen}]{yang2018breaking}
Yang, Z.; Dai, Z.; Salakhutdinov, R.; and Cohen, W.~W. 2018{\natexlab{b}}.
\newblock Breaking the softmax bottleneck: A high-rank RNN language model.
\newblock In \emph{ICLR}.

\bibitem[{Yu and Dredze(2015)}]{fct}
Yu, M.; and Dredze, M. 2015.
\newblock Learning composition models for phrase embeddings.
\newblock \emph{Transactions of the Association for Computational Linguistics}
  3: 227--242.

\bibitem[{Zheng and Lapata(2019)}]{zheng2019sentence}
Zheng, H.; and Lapata, M. 2019.
\newblock Sentence Centrality Revisited for Unsupervised Summarization.
\newblock In \emph{ACL}.

\end{thebibliography}

\newpage
\appendix



\section{Structure of Appendix}
We conduct more comprehensive experiments and analyses in Section~\ref{sec:more_exp}. The details of our method and experiments (e.g., training algorithm, preprocessing, and hyperparameter settings) are presented in Section~\ref{sec:exp_details}, and we visualize more codebook embeddings and the derived attention weights of the sentences in Section~\ref{sec:more_examples}.


\section{More Experiments}
\label{sec:more_exp}
In the main paper, we show that multi-facet embeddings can improve the estimation of symmetric relations like similarity. To know whether they are also useful in asymmetric relations like entailment, we test our method on a hypernym detection dataset in Section~\ref{sec:hyper}.

Due to the page limits, we cannot present all of our results in the main paper, so we put more comprehensive analyses for sentence similarity tasks in Section~\ref{sec:sent_sim_more}, for extractive summarization in Section~\ref{sec:sum_more}, and for phrase similarity tasks in Section~\ref{sec:phrase_sim_more}. We also present the results of BERT Large model in Section~\ref{sec:bert_large} as a reference. Section~\ref{sec:good_example_STS} and~\ref{sec:good_example_sum} provide some motivating examples for a sentence similarity task and for the extractive summarization, respectively.


\subsection{Unsupervised Hypernymy Detection}
\label{sec:hyper}
We apply our model to HypeNet~\citep{hypenet}, an unsupervised hypernymy detection dataset, based on the assumption that the co-occurring words of a phrase are often less related to some of its hyponyms. For instance, \emph{animal} is a hypernym of \emph{brown dog}. \emph{flies} is a co-occurring word of \emph{animal} which is less related to \emph{brown dog}. 



Accordingly, the predicted codebook embeddings of a hyponym $S^{hypo}_q$ (e.g., \emph{brown dog}), which cluster the embeddings of co-occurring words (e.g., \emph{eats}), often reconstruct the embeddings of its hypernym $S^{hyper}_q$ (e.g., \emph{animal}) better than the other way around (e.g., the embedding of \emph{flies} cannot reconstruct the embeddings of \emph{brown dog} well).  That is, $Er(\bm{\widehat{F}_u(S^{hypo}_q)}, \bm{W(S^{hyper}_q)})$ is smaller than $Er(\bm{\widehat{F}_u(S^{hyper}_q)}, \bm{W(S^{hypo}_q)})$).

Based on the assumption, our asymmetric scoring function is defined as

\vspace{-4mm}
\footnotesize
\begin{align}
 & \text{Diff}(S^{hyper}_q, S^{hypo}_q)  = Er(\bm{\widehat{F}_u(S^{hyper}_q)}, \bm{W(S^{hypo}_q)}) \nonumber \\
 & - Er(\bm{\widehat{F}_u(S^{hypo}_q)}, \bm{W(S^{hyper}_q)}).
 \end{align}
\normalsize
where Er function is defined in~\refeqerdot



The AUC of detecting hypernym among other relations and accuracy of detecting the hypernym direction are compared in Table~\ref{tb:phrase_hyper}. Our methods outperform baselines, which only provide symmetric similarity measurement, and \textbf{Ours (K=1)} performs similarly compared with \textbf{Ours (K=10)}.

\begin{table}[t!]
\scalebox{1}{
\begin{tabular}{|c|c|cc|cc|}
\hline
\multicolumn{2}{|c|}{Method} & \multicolumn{2}{|c|}{Dev} & \multicolumn{2}{|c|}{Test}  \\ \hline
Model & Score& AUC & Acc  & AUC & Acc \\ \hline
\multirow{2}{*}{BERT} & CLS & 20.6 & 50 & 21.3 & 50 \\ 
& Avg & 25.6 & 50 & 25.6 & 50 \\ \hline
GloVe & Avg & 17.4 & 50 & 17.7 & 50 \\ 
Our c K10 & Diff & \textbf{29.4} & 78.9  & \textbf{29.6} & 79.1 \\ 
Our c K1 & Diff & 29.3 & \textbf{82.7} & \textbf{29.6}  & \textbf{81.0} \\ \hline
\end{tabular}
}
\centering
\caption{Hypernym detection performances in the development and test set of HypeNet. AUC (\%) refers to the area under precision and recall curve, which measures the quality of retrieving hypernym phrases. Acc (\%) means the accuracy of predicting specificity given a pair of hypernym phrases.}
\label{tb:phrase_hyper}
\end{table}

\begin{table}[t!]
\scalebox{0.8}{
\begin{tabular}{|cc|cc|cc|}
\hline
\multicolumn{2}{|c|}{Method} & \multicolumn{2}{|c|}{Dev} & \multicolumn{2}{|c|}{Test} \\ \hline
\multicolumn{1}{|c|}{Score} & Model & All & Low & All & Low   \\ \hline
\multicolumn{1}{|c|}{Cosine} & Skip-thought & 43.2 & 28.1 & 30.4 & 21.2 \\ \hline
\multicolumn{1}{|c|}{Avg} & ELMo & 65.6 &	47.4 &	54.2&	44.1 \\ \hline
\multirow{3}{*}{Prob\_avg} & \multicolumn{1}{|c|}{ELMo} & 70.3 &	54.6 &	60.4 & 54.2 \\ 
 & \multicolumn{1}{|c|}{Our a (GloVe) K1} & 69.3 &	54.1 & 60.8 & 55.8 \\
 & \multicolumn{1}{|c|}{Our a (GloVe) K10} & 70.5 &	55.9 &	61.1 & 56.6 \\ \hline
\multicolumn{1}{|c|}{Avg} & \multirow{2}{*}{BERT} & 62.3 & 42.1 & 51.2 & 39.1 \\ 
\multicolumn{1}{|c|}{Prob\_avg} &  & 72.1 & 57.0 & 57.8 & 55.1 \\ \hline
\multirow{3}{*}{Avg} & \multicolumn{1}{|c|}{Sent2Vec} & 71.9 & 51.2 & 63.6 & 46.0 \\
& \multicolumn{1}{|c|}{Our a (GloVe) K10} & 76.1 & 62.9 & \textbf{71.5} & \textbf{62.7} \\
& \multicolumn{1}{|c|}{Our a (GloVe) K1} & 72.0 & 56.1 & 66.8 & 55.7 \\ \hline
\multirow{5}{*}{SC} & \multicolumn{1}{|c|}{NNSC clustering K10} & 38.6&	37.8&	25.4&	38.9 \\
 & \multicolumn{1}{|c|}{Our c (w2v) K10} & 54.7&	38.8&	43.9&	36.0 \\ 
  & \multicolumn{1}{|c|}{Our c (k-means) K10} & 37.8&	25.9&	29.5&	19.7 \\ 
  & \multicolumn{1}{|c|}{Our c (LSTM) K10} & 58.9 & 49.2 & 49.8 & 46.4 \\ 
  & \multicolumn{1}{|c|}{Our c (GloVe) K10} & 63.0 & 51.8 & 52.6 & 47.8 \\ \hline
\multirow{2}{*}{Prob\_WMD} & \multicolumn{1}{|c|}{w2v} & 72.9&	56.6&	62.1&	54.0 \\
 & \multicolumn{1}{|c|}{Our a (w2v) K10} & 73.6&	60.1&	63.5&	57.8 \\ \hline
 \multirow{2}{*}{Prob\_avg} & \multicolumn{1}{|c|}{w2v} & 68.3&	53.7&	54.3&	50.9 \\
 & \multicolumn{1}{|c|}{Our a (w2v) K10} & 68.3&	56.8&	55.1&	53.1 \\ \hline
 \multirow{2}{*}{SIF$\dagger$} & \multicolumn{1}{|c|}{w2v} & 70.5&	56.9&	59.4&	54.7 \\
 & \multicolumn{1}{|c|}{Our a (w2v) K10} & 71.6&	60.9&	61.3&	57.6 \\ \hline
\multirow{4}{*}{Prob\_WMD} & \multicolumn{1}{|c|}{GloVe} & 75.1 & 59.6 &	63.1 &	52.5 \\
  & \multicolumn{1}{|c|}{Our a (k-means) K10} & 72.5&	57.9&	60.3&	49.9 \\ 
 & \multicolumn{1}{|c|}{Our a (LSTM) K10} & \textbf{76.3}&	\textbf{63.2}&	65.8&	57.4 \\ 
 & \multicolumn{1}{|c|}{Our a (GloVe) K10} & 76.2 & 62.6 & 66.1 & 58.1 \\ \hline
\multirow{4}{*}{Prob\_avg} & \multicolumn{1}{|c|}{GloVe} & 70.7 & 56.6 & 59.2	& 54.8 \\
  & \multicolumn{1}{|c|}{Our a (k-means) K10} & 66.6&	53.4&	55.8&	51.8 \\ 
 & \multicolumn{1}{|c|}{Our a (LSTM) K10} & 71.7 & 60.1 & 61.3 & 58.3 \\ 
 & \multicolumn{1}{|c|}{Our a (GloVe) K10} & 72.0 & 60.5 & 61.4 & 59.3 \\ \hline
 \multirow{4}{*}{SIF$\dagger$} & \multicolumn{1}{|c|}{GloVe} & 75.1 & 65.7 & 63.2 & 58.1 \\
  & \multicolumn{1}{|c|}{Our a (k-means) K10} & 71.5&	62.3&	61.5&	57.2 \\ 
 & \multicolumn{1}{|c|}{Our a (LSTM) K10} & 74.6 & 66.9 & 64.3 & 60.9 \\ 
 & \multicolumn{1}{|c|}{Our a (GloVe) K10} &\textbf{75.2} & \textbf{67.6} & \textbf{64.6} & \textbf{62.2} \\ \hline
\end{tabular}
}
\centering
\caption{The Pearson correlation (\%) in STS benchmarks. w2v means Word2Vec. Our * (k-means) means using the k-means loss rather than the NNSC loss. Our * (LSTM) means replacing the transformers in our encoder with bi-LSTM and replacing our transformer decoder with LSTM. Other abbreviations and symbols share the same meaning in Table~\reftbsts.}
\label{tb:STS_more}
\end{table}

\begin{table*}[t!]
\scalebox{0.8}{
\begin{tabular}{|ccc|ccc||ccc||cccc|}
\hline
\multicolumn{3}{|c|}{Dataset} & \multicolumn{3}{c||}{Prob\_avg} & \multicolumn{2}{|c|}{Prob\_WMD} & WMD & \multicolumn{4}{|c|}{Avg} \\ \hline
Category & Data & Year & Our a K10 & GloVe & Our a K1 & Our a K10 & GloVe & GloVe & w2v & BERT & ELMo & ST \\ \hline
\multirow{5}{*}{forum} &	deft-forum & 2014 & \textbf{40.3}&	33.2&	35.6&	\textbf{41.4}&	33.3&	28.5&	33.9&	25.6&	\textbf{35.9}&	22.5 \\
 & answers-forums&2015&\textbf{53.9}&49.6&43.4&\textbf{63.7}&61.4&48.7&51.2&55.1&\textbf{56.0}&36.8\\
& answer-answer&2016&\textbf{39.5}&34.1&33.1&\textbf{48.9}&45.0&46.4&31.0&50.4&\textbf{58.2}&32.3\\
& question-question&2016&\textbf{63.5}&61.4&60.1&\textbf{66.9}&62.6&28.7&\textbf{53.7}&44.9&38.4&42.3\\
& belief&2015&\textbf{58.6}&57.2&53.5&\textbf{67.8}&66.9&62.6&63.8&62.8&\textbf{71.1}&38.5\\ \hline
\multirow{5}{*}{news} & surprise.SMTnews&2012&\textbf{56.7}&50.9&54.4&\textbf{55.9}&49.0&48.4&48.7&\textbf{56.8}&55.1&44.2\\
& headlines&2013&\textbf{57.9}&55.9&53.3&\textbf{67.5}&66.2&59.6&54.8&55.1&\textbf{59.7}&44.6\\
& headlines&2014&\textbf{53.0}&51.2&49.1&\textbf{61.8}&59.9&53.3&50.4&53.2&\textbf{54.2}&41.6\\
& headlines&2015&\textbf{54.7}&51.8&49.8&\textbf{65.4}&63.6&59.1&51.8&53.3&\textbf{59.2}&47.6\\
& headlines&2016&\textbf{53.9}&52.0&49.8&\textbf{65.6}&64.6&62.6&50.3&\textbf{58.0}&57.3&46.6\\ \hline
\multirow{4}{*}{definition} & surprise.OnWN&2012&\textbf{66.6}&64.0&63.6&\textbf{69.8}&68.2&65.3&63.3&60.3&\textbf{68.8}&33.5 \\
& OnWN&2013&\textbf{69.8}&68.6&63.6&\textbf{65.0}&61.8&35.6&\textbf{63.8}&59.7&44.8&21.1 \\
& OnWN&2014&\textbf{75.9}&74.8&72.7&\textbf{73.8}&71.9&52.5&\textbf{74.5}&71.0&61.1&31.1 \\
& FNWN&2013&\textbf{39.4}&38.6&38.9&45.7&\textbf{46.0}&40.1&28.0&36.5&\textbf{38.1}&11.0 \\ \hline
\multirow{3}{*}{captions} & MSRvid&2012&\textbf{81.5}&81.2&80.5&\textbf{80.4}&78.3&46.3&\textbf{76.0}&52.3&63.0&54.3 \\
&images&2014&\textbf{76.9}&74.6&75.5&\textbf{78.1}&75.1&57.3&\textbf{72.1}&55.4&64.9&62.8 \\
&images&2015&76.1&\textbf{76.5}&74.5&\textbf{82.2}&81.8&67.3&72.8&66.3&\textbf{73.3}&29.5 \\ \hline
\multirow{2}{*}{education} & answers-students&2015&53.3&\textbf{54.7}&52.5&66.4&68.5&\textbf{70.4}&\textbf{64.0}&62.8&60.4&41.5 \\
 & plagiarism&2016&72.1&\textbf{74.4}&72.2&78.5&\textbf{79.1}&71.2&74.0&76.6&\textbf{78.4}&53.6 \\ \hline
\multirow{2}{*}{out of domain} & deft-news&2014&62.4&\textbf{65.6}&59.0&63.9&\textbf{65.2}&55.5&58.9&\textbf{73.3}&72.6&43.8 \\
& tweet-news&2014&64.0&\textbf{66.2}&60.1&71.1&\textbf{72.7}&70.5&69.7&66.5&\textbf{72.0}&53.1 \\ \hline
\multirow{3}{*}{similar} & MSRpar&2012&34.6&\textbf{43.4}&36.2&48.0&\textbf{53.2}&49.0&37.2&\textbf{40.5}&34.0&24.5 \\
 & SMTeuroparl&2012&52.3&\textbf{54.4}&42.8&\textbf{53.6}&53.3&51.3&\textbf{51.8}&46.0&46.8&28.4 \\
 & postediting&2016&65.4&\textbf{66.8}&63.5&78.9&\textbf{79.9}&78.0&74.3&79.1&\textbf{80.2}&57.4 \\ \hline
\multirow{12}{*}{STS} & All & 2012&68.4&\textbf{68.5}&68.0&\textbf{66.8}&64.9&40.0&\textbf{60.5}&32.5&44.1&6.6 \\
& All & 2013&\textbf{64.1}&62.0&58.9&\textbf{66.5}&64.2&47.3&\textbf{58.5}&57.4&54.1&36.3 \\
& All&2014&\textbf{58.6}&56.1&52.7&\textbf{61.4}&58.8&39.2&\textbf{56.1}&53.6&52.4&25.0 \\
& All&2015&\textbf{59.3}&56.8&54.2&\textbf{71.1}&70.1&58.1&\textbf{61.6}&58.1&57.4&25.8 \\
& All&2016&\textbf{56.9}&53.7&52.4&\textbf{66.7}&62.8&49.6&52.9&\textbf{60.4}&57.9&38.8 \\
& All&All&\textbf{61.4}&59.5&56.5&\textbf{66.8}&64.8&48.4&\textbf{59.1}&54.1&55.4&26.1 \\
& Low&2012&67.1&\textbf{67.9}&\textbf{67.9}&\textbf{60.8}&59.2&17.5&\textbf{57.5}&18.8&33.0&-2.0 \\
& Low&2013&\textbf{49.4}&45.1&44.3&\textbf{39.4}&32.7&10.5&\textbf{33.1}&31.5&27.4&22.9 \\
& Low&2014&\textbf{50.2}&45.1&43.5&\textbf{50.5}&45.5&22.3&43.5&\textbf{46.6}&40.4&29.4 \\ 
& Low&2015&\textbf{48.8}&45.3&43.8&\textbf{54.2}&50.9&33.4&\textbf{49.8}&38.2&41.3&18.7 \\ 
& Low&2016&\textbf{51.0}&45.1&44.5&\textbf{53.8}&46.3&21.7&\textbf{39.4}&36.4&32.4&21.6 \\
& Low&All&\textbf{51.2}&47.7&45.9&\textbf{52.3}&48.4&25.8&\textbf{45.9}&39.8&39.5&21.7 \\ \hline
\end{tabular}
}
\centering
\caption{Comparing Pearson correlation (\%) of different unsupervised methods from STS 2012 to STS 2016. We highlight the best performance in each of the three blocks.}
\label{tb:STS_all}
\end{table*}

\subsection{More Analysis on Sentence Similarity }
\label{sec:sent_sim_more}
We design more experiments and present the results in Table~\ref{tb:STS_more} and Table~\ref{tb:STS_all} in order to answer the following research questions.

\textbf{1. Is ignoring the order of co-occurring words effective in emphasizing the semantic side of the sentences?}

To answer this question, we replace our transformer encoder with bi-LSTM and our transformer decoder with LSTM. Then, this architecture becomes very similar to skip-thought~\citep{kiros2015skip} except that skip-thoughts decodes a sequence instead of a set, and we ignore the word order in the nearby sentences. As we can see in Table~\ref{tb:STS_more}, \textbf{Our c (LSTM) K10 SC} performs much better than \textbf{Skip-thought Cosine}, which compute the cosine similarity between their sentence embeddings. This result further justifies our approach of ignoring the order of co-occurring words in our NNSC loss.





\textbf{2. Is our word importance estimation generally useful for composing (contextualized) word embedding models?}

We cannot apply our attention weights (i.e.,  \textbf{Our a}) to BERT because BERT uses word piece tokenization. Instead, we use the top layer of ELMo~\citep{ELMo} as the contextualized word embedding, apply $\frac{\alpha}{\alpha + p(w)}$ weighting multiplied with our attention weights in~\refeqatt. The results in Table~\ref{tb:STS_more} show that the performance of \textbf{ELMo Prob\_avg} could also be boosted by our attention weighting even though our model is trained on GloVe semantic space. The importance weights from multiple embeddings can also help boost the performance of a version of Sent2Vec~\citep{pgj2017unsup} that uses only unigram information.

\textbf{3. Could our model be trained on word embedding space other than GloVe?}

First, we train Word2Vec~\citep{word2vec} (denoted as w2v) on the Wikipedia 2016 corpus. We then train our multi-facet embeddings to fit the Word2Vec embedding of co-occurring words in the Wikipedia 2016 corpus. The results in Table~\ref{tb:STS_more} show that \textbf{Our a (w2v) K10} improves the performance using different scoring functions as we did in GloVe space.

\textbf{4. How well could clustering-based multi-facet embeddings perform on long text sequences such as sentences?}

Lots of the testing sentences in the STS benchmark are not observed in our training corpus. To test clustering-based multi-facet embeddings, we first average word embedding in every sentence into sentence embedding, and for each testing query sentence, we perform approximated nearest neighbor search using KDTree~\citep{bentley1975multidimensional} to retrieve 1000 most similar sentences. Then, we remove the stop words in the 1000 sentences and perform NNSC clustering on the rest of the words. Finally, we compute \textbf{SC} distance between two sets of cluster centers derived from testing sentence pairs and denote the baseline as \textbf{NNSC clustering K10 SC} in Table~\ref{tb:STS_more}. 

The testing time of this baseline is much slower than the proposed method due to the need for the nearest neighbor search, and its performance is also much worse. This result justifies our approach of predicting clustering centers directly to generate multi-facet embeddings.

\textbf{5. How much better is NNSC loss compared with k-means loss?}

In the method section, we mention that we adopt NNSC rather than k-means in our loss because k-means loss cannot generate diverse cluster centers in all of the neural architectures (including transformers and bi-LSTMs) we tried. We hypothesize that the k-means loss does not stably encourage predicted clusters to play different roles for reconstructing the embeddings of observed co-occurring words. We present the much worse results of the model using k-means loss in Table~\ref{tb:STS_more} to justify our usage of NNSC in our loss.


\textbf{6. Could our method improve the similarity estimation of all kinds of datasets?}

In Table~\ref{tb:STS_all}, we compare the performance before and after applying our attention weights in the English part of STS 2012~\citep{agirre2012semeval}, 2013~\citep{agirre2013sem}, 2014~\citep{agirre2014semeval}, 2015~\citep{agirre2015semeval}, and 2016~\citep{agirre2016semeval}. We categorize each of the dataset in different years based on either its source (\emph{forum}, \emph{news}, \emph{definition}, \emph{caption}, and \emph{education}) or its characteristic (\emph{out of domain} or \emph{similar}). 

\emph{Out of domain} means the testing sentences are very different from our training corpus, Wikipedia 2016. \emph{deft-news} from STS 2014 is included in this category because all the sentences in the dataset are lowercased. \emph{Similar} means there are lots of pairs in the datasets whose two sentences have almost the identical meaning. 

From the Table~\ref{tb:STS_all}, we can see that \textbf{GloVe Prob\_avg} and \textbf{GloVe Prob\_WMD} perform well compare with other baselines, and the attention weights from our multi-facet embedding stably boost \textbf{GloVe Prob\_avg} and \textbf{GloVe Prob\_WMD} except in the categories \emph{education}, \emph{out of domain}, and \emph{similar}. Thus, we recommend adopting our method when the source of training and testing sentences are not too different from each other, and the task is not to identify duplicated sentences.

\textbf{7. Are supervised methods such as sentence-BERT sensitive to the training data?}

Table~\ref{tb:sentence_bert_details} compares the performance of sentence-BERT~\citep{reimers2019sentence} trained on different data sources. We observe that the performance of sentence-BERT could be degraded when the distribution of training data is very different from that of testing data. For example, Sentence-BERT also does not perform well when the training sentence pairs tend to be similar with each other (e.g., in \emph{postediting} and \emph{SMTeuroparl}) or come from a writing style that is different from the style of testing sentence pairs (e.g., \emph{tweet-news} and \emph{answers-students}). 

Furthermore, a supervised model trained by a limited amount of labels could perform worse than the unsupervised alternatives. For example, on STSB Dev, the weighted average of word embedding (Prob\_avg) outputted by the BERT base model outperforms the sentence-BERT trained by 100 labels on average. Sentence-BERT model trained by \emph{SMTeuroparl} is even worse than just averaging all the contextualized word embeddings in BERT on STSB Test.

\begin{table}[t!]
\scalebox{0.9}{
\begin{tabular}{|cc|cc|cc|}
\hline
\multicolumn{2}{|c|}{Training Data} & \multicolumn{2}{|c|}{Dev} & \multicolumn{2}{|c|}{Test} \\ \hline
\multicolumn{1}{|c|}{Data Source} & Year & All & Low & All & Low   \\ \hline

\multicolumn{1}{|c|}{SMTeuroparl}&	2012&63.5&41.7&49.6&43.3 \\
\multicolumn{1}{|c|}{surprise.SMTnews}&	2012&67.1&49.2&58.4&56.5 \\
\multicolumn{1}{|c|}{postediting}&	2016&70.3&53.9&60.5&56.3 \\
\multicolumn{1}{|c|}{tweet-news}&	2014&68.8&52.4&61.4&57.1 \\
\multicolumn{1}{|c|}{answers-students}&	2015&67.5&53.1&62.5&56.4 \\
\multicolumn{1}{|c|}{headlines}&	2014&69.2&52.8&62.6&53.8 \\
\multicolumn{1}{|c|}{plagiarism}&	2016&72.1&59.5&65.4&62.4 \\
\multicolumn{1}{|c|}{belief}&	2015&71.9&53.5&65.7&59.1 \\
\multicolumn{1}{|c|}{FNWN}&	2013&71.2&54.7&67.1&61.9 \\
\multicolumn{1}{|c|}{headlines}&	2015&73.8&58.9&67.9&53.1 \\
\multicolumn{1}{|c|}{question-question}&	2016&75.0&63.2&69.3&65.5 \\
\multicolumn{1}{|c|}{OnWN}&	2013&\textbf{75.9}&63.4&70.3&64.0 \\
\multicolumn{1}{|c|}{OnWN}&	2014&74.9&\textbf{63.5}&70.6&\textbf{65.6} \\
\multicolumn{1}{|c|}{surprise.OnWN}&	2012&75.5&57.5&\textbf{71.5}&60.6 \\ \hline
\multicolumn{2}{|c|}{Average}	&71.2&55.5&64.5&58.2 \\ \hline

\end{tabular}
}
\centering
\caption{The Pearson correlation (\%) of sentence-BERT on STS benchmark. The sentence-BERT is initialized by the BERT base model and trained by 100 samples in each data source. All results are the average of three runs. The order of rows is determined by their performance on the test set of STSB.}
\label{tb:sentence_bert_details}
\end{table}



\subsection{Summarization Comparison Given the Same Summary Length}
\label{sec:sum_more}

\begin{figure}[t!]
\begin{center}
\includegraphics[width=1\linewidth]{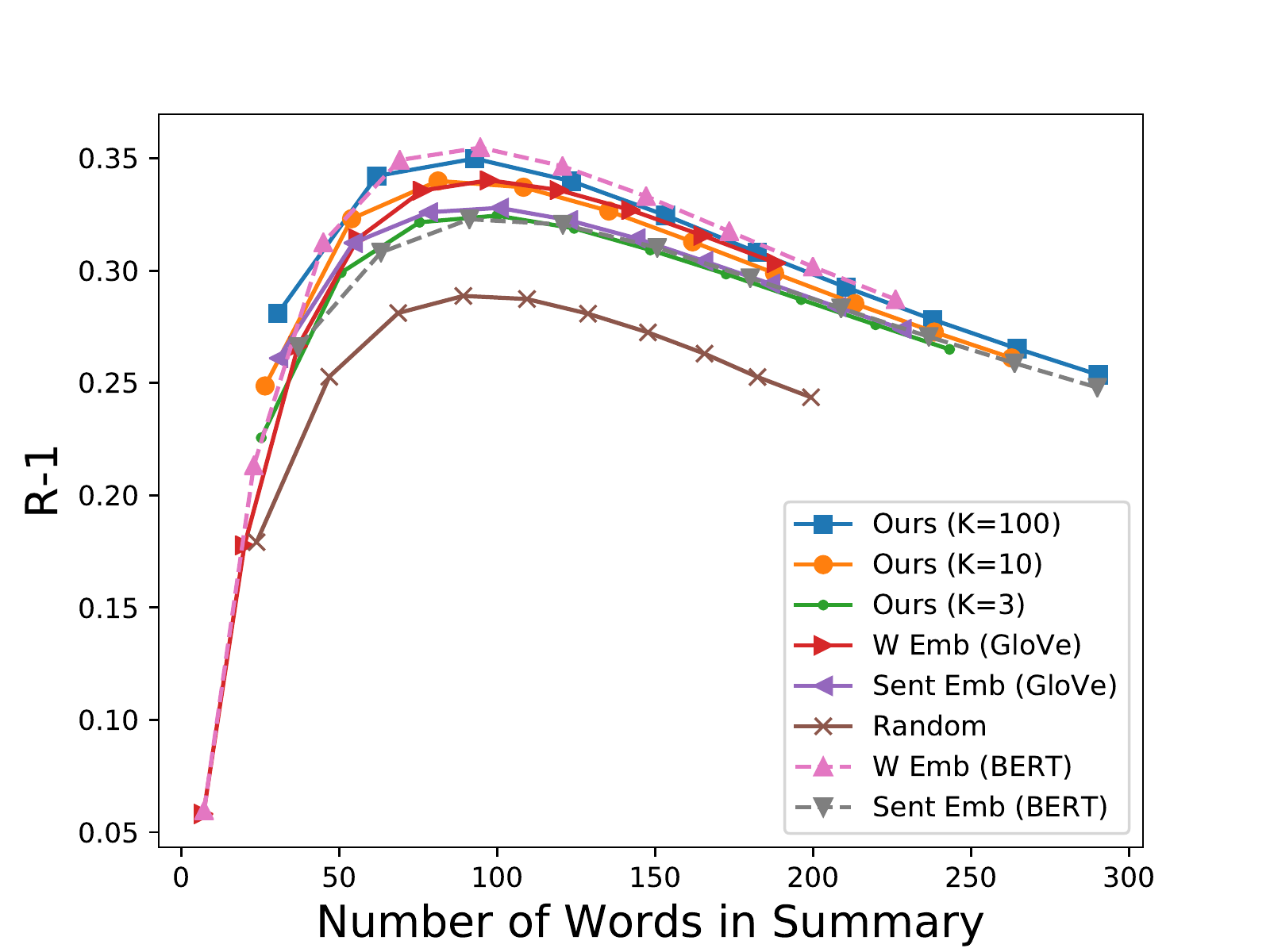}
\end{center}
\caption{Comparing the F1 of ROUGE-1 score on unsupervised methods that do not access the sentence order information in CNN/Daily Mail.}
\label{fig:summarization}
\end{figure}

In Section~\refsum
, we compare our methods with other baselines when all the methods choose the same number of sentences. We suspect that the bad performances for \textbf{W Emb (*)} methods (i.e., representing each sentence using the embedding of words in the sentence) might come from the tendency of selecting shorter sentences.
To verify the hypothesis, we plot the R-1 performance of different unsupervised summarization methods that do not use the sentence order information versus the sentence length in Figure~\ref{fig:summarization}. 

In the figure, we first observe that \textbf{Ours (K=100)} significantly outperforms \textbf{W Emb (GloVe)} and \textbf{Sent Emb (GloVe)} when summaries have similar length. In addition, we find that \textbf{W Emb (*)} usually outperforms \textbf{Sent Emb (*)} when comparing the summaries with a similar length. Notice that this comparison might not be fair because \textbf{W Emb (*)} are allowed to select more sentences given the same length of summary and it might be easier to cover more topics in the document using more sentences. In practice, preventing choosing many short sentences might be preferable in an extractive summarization if fluency is an important factor.

Nevertheless, suppose our goal is simply to maximize the ROUGE F1 score given a fixed length of the summary without accessing the ground truth summary and sentence order information. In that case, the figure indicates that \textbf{Ours (K=100)} significantly outperform \textbf{W Emb (GloVe)} and is the best choice when the summary length is less than around 50 words and \textbf{W Emb (BERT)} becomes the best method for a longer summary. The BERT in this figure is the BERT base model. The mixed results suggest that combining our method with BERT might be a promising direction to get the best performance in this task (e.g., use contextualized word embedding from BERT as our pre-trained word embedding space).

\begin{table*}[t!]
\scalebox{1}{
\begin{tabular}{|cc|c|cc|cc|}
\hline
 & & \multicolumn{3}{|c|}{Lowercase} & \multicolumn{2}{|c|}{Uppercase}  \\ \hline 
\multicolumn{2}{|c|}{Method} & BiRD & WikiSRS-Sim & WikiSRS-Rel & WikiSRS-Sim & WikiSRS-Rel  \\ \hline 
\multicolumn{1}{|c|}{Model} & Score & Pearson & \multicolumn{4}{|c|}{Spearman}  \\ \hline
ELMo & \multicolumn{1}{|c|}{Avg} &  - & - & - & 54.4 &	49.2  \\ 
BERT & \multicolumn{1}{|c|}{Avg} &  44.4 & 43.3 & 40.7 & 40.3 & 37.8  \\ 
GloVe & \multicolumn{1}{|c|}{Avg}  & 47.9\footnotemark & 47.7 & 49.1 & 62.7 &	62.7 \\ 
Ours (K=10) & \multicolumn{1}{|c|}{Emb} & 57.3 & \textbf{67.4} & \textbf{66.9} & 68.6 & \textbf{69.6} \\ 
Ours (K=1) & \multicolumn{1}{|c|}{Emb} & \textbf{60.6} & 67.1 & 65.8 & \textbf{69.4} &	69.4 \\ \hline
\end{tabular}
}
\centering
\caption{Performances of phrase similarity tasks. In BiRD and WikiSRS, the correlation coefficient (\%) between the predicted similarity and the ground truth similarity is presented. }
\label{tb:phrase_sim_more}
\end{table*}

\begin{table}[t!]
\scalebox{0.8}{
\begin{tabular}{|cc|c|c|}
\hline
\multicolumn{3}{|c|}{Similarity} & Summarization  \\ \hline
\multicolumn{2}{|c|}{STSB All} & STS2012-6 All & CNN DM  \\	
Dev & Test & Test& Dev   \\
1,500 & 1,379 & 12,544 & 11,490  \\ \hline
\end{tabular}
}
\centering
\caption{Dataset sizes for sentence representations.}
\label{tb:dataset_size_sent}
\end{table}

\begin{table}[t!]
\scalebox{0.8}{
\begin{tabular}{|c|c|c|cc|cc|}
\hline
\multicolumn{5}{|c|}{Similarity} & \multicolumn{2}{|c|}{Hypernym} \\ \hline
SemEval 2013 & Turney2012 & BiRD & \multicolumn{2}{|c|}{WikiSRS} & \multicolumn{2}{|c|}{HypeNet}	\\
Test & Test & Test &Sim & Rel & Val & Test \\
7,814 & 1,500 & 3,345 &688 & 688 & 3,534 & 17,670 \\ \hline
\end{tabular}
}
\centering
\caption{Dataset sizes for phrase representations.}
\label{tb:dataset_size_phrase}
\end{table}

\subsection{Experiments on More Phrase Similarity Datasets}
\label{sec:phrase_sim_more}
We conduct the phrase similarity experiments on two recently proposed datasets, BiRD~\citep{bird-naacl2019}, and WikiSRS~\citep{WikiSRS}, which contain ground truth phrase similarities derived from human annotations. BiRD and WikiSRS-Rel measure the relatedness of phrases and WikiSRS-Sim measures the similarity of phrases. The phrases are proper nouns in WikiSRS and are mostly common nouns in BiRD. Since the main goal of WikiSRS is to test the entity representation, we also test the different models trained on the corpus without lowercasing all the words. 

The results are presented in Table~\ref{tb:phrase_sim_more}. The multi-facet embedding performs similarly compared with single-facet embedding and is better than other baselines. This result confirms our findings in the main paper that the phrase similarity performance is not sensitive to the number of clusters $K$.

\begin{table}[t!]
\centering
\scalebox{0.9}{
\begin{tabular}{|c|c|c|c|}
\hline
Method & Hidden size & \#Parameters & Testing Time \\ \hline
K=1 & 300 & 6.7M & 9 ms \\
K=10 & 300 & 13.7M & 18 ms \\
BERT Base & 768 & 86.0M & 18 ms \\
BERT Large & 1024 & 303.9M & 65 ms \\ \hline
\end{tabular}
}
\centering
\caption{Comparison of model sizes. The number of parameters does not include the word embedding layer. We show the test time required for a batch with 50 sentences using one 1080Ti GPU.}
\label{tb:model_size}
\end{table}

\begin{table*}[t!]
\centering
\begin{minipage}{.5\linewidth}
\centering
\scalebox{0.9}{
\begin{tabular}{|cc|cc|cc|}
\hline
\multicolumn{2}{|c|}{Method} & \multicolumn{2}{|c|}{Dev} & \multicolumn{2}{|c|}{Test} \\ \hline
\multicolumn{1}{|c|}{Model} & Score & All & Low & All & Low   \\ \hline
BERT Base & \multicolumn{1}{|c|}{Prob\_avg} &72.1 & 57.0 & 57.8 & 55.1 \\ 
BERT Large & \multicolumn{1}{|c|}{Prob\_avg} & 74.3 & 61.0 & 65.0 & \textbf{60.0} \\ \hline
\multirow{2}{*}{Our a (GloVe) K10} & \multicolumn{1}{|c|}{Prob\_avg} & 72.0 & 60.5 & 61.4 & 59.3 \\ 
& \multicolumn{1}{|c|}{Prob\_WMD} & \textbf{76.2} & \textbf{62.6} & \textbf{66.1} & 58.1 \\ \hline
\end{tabular}
}
\caption{Compare BERT Large with Ours in Table~\reftbsts
.}
\label{tb:STS_bert}
\end{minipage}%
\quad 
\begin{minipage}{.48\linewidth}
\centering
\scalebox{1}{
\begin{tabular}{|c|c|c|c|c|}
\hline
\multicolumn{2}{|c|}{Method} & R-1 & R-2 & Len \\ \hline
\multirow{2}{*}{BERT Base} & W Emb & 31.2 & 11.2 & 44.9 \\ 
 & Sent Emb  & 32.3  & 10.6 & 91.2 \\ \hline 
\multirow{2}{*}{BERT Large} & W Emb & 31.1 &	11.0 & 46.8 \\ 
 & Sent Emb  & 32.7 & 10.9 & 86.5 \\ \hline
Our c (K=100) & Bases & \textbf{35.0} & \textbf{12.8} & 92.9 \\ \hline

\end{tabular}
}
\caption{Compare BERT Large with Ours in Table~\reftbsum .}
\label{tb:sum_bert}
\end{minipage}
\end{table*}

\begin{table*}[t!]
\scalebox{0.8}{
\begin{tabular}{|cc|cc|c|c|c|c|c|c|c|}
\hline
& & \multicolumn{7}{|c|}{Lowercased} & \multicolumn{2}{|c|}{Uppercased} \\ \cline{3-11}
\multicolumn{2}{|c|}{Method} & \multicolumn{2}{c|}{SemEval} & Turney (5) & Turney (10) & BiRD & WikiSRS-Sim & WikiSRS-Rel & WikiSRS-Sim & WikiSRS-Rel  \\ \hline 
\multicolumn{1}{|c|}{Model} & Score & AUC & F1 & Acc & Acc & Pearson & \multicolumn{4}{|c|}{Spearman}  \\ \hline
BERT Base & \multicolumn{1}{|c|}{Avg} & 66.5 & 67.1 & 43.4 & 24.3 & 44.4 & 43.3 & 40.7 & 40.3 &	37.8  \\ 
BERT Large & \multicolumn{1}{|c|}{Avg} & 72.4 & 66.7 & \textbf{51.3} & 32.1 & 47.5 & 49.6 & 48.1 & 28.6 & 34.0  \\ 
Ours (K=1) & \multicolumn{1}{|c|}{Emb} & \textbf{87.8} & \textbf{78.6} & 50.3 & \textbf{32.5} & \textbf{60.6} & \textbf{67.1} & \textbf{65.8} & \textbf{69.4} &	\textbf{69.4} \\ \hline
\end{tabular}
}
\centering
\caption{Compare BERT Large with Ours in Table~\reftbphrase
. }
\label{tb:phrase_sim_bert}
\end{table*}

\begin{table}[t!]
\centering
\scalebox{1}{
\begin{tabular}{|c|cc|cc|}
\hline
\multirow{2}{*}{Method} & \multicolumn{2}{|c|}{Dev} & \multicolumn{2}{|c|}{Test}  \\ \cline{2-5}
& AUC & Acc  & AUC & Acc \\ \hline
BERT Base (Avg) & 25.6 & 50 & 25.6 & 50 \\ 
BERT Large (Avg) & 20.2 & 50  & 20.1  & 50 \\ 
Ours (K=1) & \textbf{29.3} & \textbf{82.7} & \textbf{29.6}  & \textbf{81.0} \\ \hline
\end{tabular}
}
\centering
\caption{Compare BERT Large with Ours in Table~\ref{tb:phrase_hyper}.}
\label{tb:phrase_hyper_bert}

\end{table}

\begin{table*}[t!]
\scalebox{0.85}{
\begin{tabular}{|cc|c|ccccc|}
\hline
\multirow{2}{*}{Sentence 1} & \multirow{2}{*}{Sentence 2} & Score & \multicolumn{5}{|c|}{Score Rank Among 1500 Pairs} \\
 &  & GT & GT & Our c & Prob\_avg + Our a  & Prob\_avg & Avg \\ \hline
A turtle walks over the ground . & A large turtle crawls in the grass . & 3.75 & 326 & 400 & 638 & 717 & 761 \\
The animal with the big eyes is eating . & A slow loris is eating . & 2.60 & 690 & 611 & 1001 & 1223 & 1370 \\
A man is riding on a horse . & A woman is riding an elephant . & 1.53 & 1021 & 869 & 722 & 549 & 540 \\ \hline
\end{tabular}
}
\centering
\caption{Motivating examples for a sentence similarity task. The sentences are image captions from MSRvid dataset in STS 2012. GT means ground truth. All our methods here set $K=10$.}
\label{tb:good_example_STS}
\end{table*}

\begin{table*}[t!]
\scalebox{0.95}{
\begin{tabular}{|c|c|c|}
\hline
& Sentence 1 & Sentence 2 \\ \hline
Sentences & \colorbox{c83}{A} \colorbox{c22}{turtle} \colorbox{c41}{walks} \colorbox{c42}{over} \colorbox{c45}{the} \colorbox{c18}{ground} \colorbox{c61}{.} & \colorbox{c82}{A} \colorbox{c36}{large} \colorbox{c32}{turtle} \colorbox{c61}{crawls} \colorbox{c60}{in} \colorbox{c52}{the} \colorbox{c28}{grass} \colorbox{c64}{.} \\ \hline
& can 0.876 you 0.846 even 0.845 & can 0.857 even 0.838 often 0.832 \\
& turtle 0.914 turtles 0.822 dolphin 0.755 & turtle 0.917 turtles 0.818 dolphin 0.755 \\
& hillside 0.737 hills 0.711 mountains 0.704 & beneath 0.697 hillside 0.645 down 0.639 \\
& species 0.937 habitat 0.771 habitats 0.759 & species 0.949 habitat 0.778 habitats 0.760  \\
Output & animals 0.689 pigs 0.675 animal 0.658 & females 0.655 males 0.630 male 0.622  \\
Embeddings& white 0.842 blue 0.839 red 0.833 & white 0.863 blue 0.862 red 0.856  \\
& insectivorous 0.650 immatures 0.627 insectivores 0.618 & immatures 0.616 foliose 0.609 tussocks 0.607  \\
& ground 0.693 soil 0.676 surface 0.626 & seawater 0.644 salinity 0.597 soils 0.593 \\
& ground 0.861 grass 0.598 Ground 0.576 & grass 0.870 grasses 0.739 weeds 0.713   \\
& waking 0.551 strolls 0.551 wander 0.546 & length 0.732 width 0.639 diameter 0.627  \\ \hline
Sentences & \colorbox{c82}{The} \colorbox{c24}{animal} \colorbox{c56}{with} \colorbox{c54}{the} \colorbox{c51}{big} \colorbox{c31}{eyes} \colorbox{c58}{is} \colorbox{c28}{eating} \colorbox{c66}
{.} & \colorbox{c84}{A} \colorbox{c47}{slow} \colorbox{c64}{loris} \colorbox{c62}{is} \colorbox{c27}{eating} \colorbox{c65}{.} \\ \hline
& even 0.887 sure 0.877 want 0.868 & often 0.880 usually 0.860 sometimes 0.838 \\
& animals 0.896 animal 0.896 rabbits 0.678 & loris 0.864 lorises 0.670 langur 0.596   \\
& food 0.825 foods 0.802 eating 0.798 & foods 0.763 food 0.709 nutritional 0.690 \\
& fingers 0.695 legs 0.692 shoulders 0.691 & eating 0.799 food 0.798 eat 0.794 \\
Output & species 0.919 habitats 0.738 habitat 0.731 & species 0.949 habitat 0.787 habitats 0.769 \\
Embeddings & blue 0.834 red 0.809 white 0.805 & blue 0.844 white 0.839 red 0.827 \\
& ingestion 0.608 inflammation 0.591 concentrations 0.588 & gently 0.649 wet 0.642 beneath 0.641\\
& male 0.652 female 0.636 disease 0.618 & male 0.685 female 0.659 females 0.658  \\
& profound 0.668 perceived 0.647 profoundly 0.634  &  decreasing 0.710 decreases 0.699 decrease 0.697  \\
& beady 0.626 beaks 0.623 mandibles 0.602 & C. 0.624 L. 0.620 A. 0.593   \\ \hline
Sentences& \colorbox{c78}{A} \colorbox{c21}{man} \colorbox{c52}{is} \colorbox{c27}{riding} \colorbox{c67}{on} \colorbox{c52}{a} \colorbox{c16}{horse} \colorbox{c66}{.} & \colorbox{c80}{A} \colorbox{c14}{woman} \colorbox{c50}{is} \colorbox{c33}{riding} \colorbox{c57}{an} \colorbox{c16}{elephant} \colorbox{c67}{.} \\ \hline
& sure 0.883 even 0.883 want 0.867 & sure 0.880 even 0.876 know 0.866 \\
& slid 0.686 legs 0.681 shoulders 0.670 & underneath 0.701 shoulders 0.671 legs 0.668  \\
& moisture 0.518 drying 0.517 coated 0.516 & elephant 0.912 elephants 0.842 hippo 0.759  \\
& horse 0.917 horses 0.878 stallion 0.728 & elephant 0.885 elephants 0.817 animals 0.728   \\
Output & mortals 0.668 fearful 0.664 beings 0.646 & fearful 0.677 disdain 0.637 anguish 0.632  \\
Embeddings & man 0.864 woman 0.764 boy 0.700 & woman 0.822 women 0.787 female 0.718 \\
& man 0.858 woman 0.652 boy 0.637 & girl 0.784 woman 0.781 lady 0.721  \\
& discovers 0.647 informs 0.638 learns 0.634 & discovers 0.662 learns 0.659 realizes 0.648 \\
& race 0.812 races 0.800 championship 0.697 & movie 0.622 film 0.615 films 0.590  \\
& Horse 0.763 Riding 0.696 Horses 0.656 & riding 0.873 bike 0.740 biking 0.731  \\ \hline
\end{tabular}
}
\centering
\caption{The predicted word importance and codebook embeddings on sentences from Table~\ref{tb:good_example_STS}. The way of visualization is the same as that in
Section~\ref{sec:more_examples}.
}
\label{tb:good_topics_STS}
\end{table*}

\begin{table*}[t!]
\scalebox{0.75}{
\begin{tabular}{|c|c|p{19cm}|}
\hline
Method & Index & \multicolumn{1}{|c|}{Selected Sentence} \\ \hline
 & NA & Swedish photographer , Erik Johansson , spends months photographing images to build up to the finished picture . \\
Ground Truth & NA & Each image is made up of hundreds of separate shots and painstakingly detailed work by the expert retoucher . \\
& NA & Erik , 30 , said : ' Can I put this very weird idea in a photograph and make it look like it was just captured ? ' \\ \hline
 & 1 & Thought the black and blue dress was an optical illusion ? \\
Lead-3 & 2 & It 's nothing compared to these mind - boggling pictures by a Swedish photographer , artist , and Photoshop genius . \\
 & 3 & Erik Johansson , 30 , who is based in Berlin , Germany , says he does n’t capture moments , but instead captures ideas . \\ \hline
 & 6 & Swedish photographer , artist , and Photoshop genius , Erik Johansson , has created mind - boggling photos like this inside - out house that look different on each glance . \\
Our c (K=10) & 42 & Reverse Opposite is mind - bendig as , with an MC Escher drawing , the car seems both on and under the bridge at the same time . \\
& 1 & Thought the black and blue dress was an optical illusion ? \\
\hline
 & 46 & Although one photo can consist of lots of different images merged into one , he always wants it to look like it could have been captured as a whole picture . \\
\multirow{2}{*}{Sent\_Emb (GloVe)} & 25 & He cites Rene Magritte , Salvador Dali and MC Escher as artistic influences . \\
 & 18 & Using Photoshop , he turned the running paint into rolling fields and superimposed a photograph of a house on to the cardboard model , adding a photo of a water wheel to complete the fantastical and dramatic shot of a dreamy , bucolic landscape that seems to be falling over a cliff . \\ \hline
 & 41 & he said . \\
W\_Emb (GloVe) & 23 & But there are tons of inspiration online . \\
& 22 & ' I think I get more inspiration from paintings rather than photos . \\ \hline
\multirow{5}{*}{Sent\_Emb (BERT)} & 18 & Using Photoshop , he turned the running paint into rolling fields and superimposed a photograph of a house on to the cardboard model , adding a photo of a water wheel to complete the fantastical and dramatic shot of a dreamy , bucolic landscape that seems to be falling over a cliff . \\
 &41 & he said . \\
&12 & He said : ' It ’s the challenge : can I put this very weird idea in a photograph and make it look like it was just captured ? ' \\ \hline
 &5 & Scroll down for video . \\
W\_Emb (BERT) & 30 & In Closing Out , interiors and exterior meld as one in this seemingly simple tableau . \\
 & 12 & He said : ' It ’s the challenge : can I put this very weird idea in a photograph and make it look like it was just captured ? ' \\ \hline
\end{tabular}
}
\centering
\caption{Motivating examples for extractive summarization. The sentences come from a document in the validation set of CNN/Daily Mail. Index indicates the sentence order in the document. Ground truth means the summary from humans. The sentences in each method are ranked by its selection order. For example, our method selects the $6$th sentence in the document first.}
\label{tb:good_example_sum}
\end{table*}

\subsection{Comparison with BERT Large}
\label{sec:bert_large}
In Table~\ref{tb:model_size}, we compare the size and running time of different models for sentence representation. As mentioned in Section~\refexpsetup
, our model has fewer parameters than the BERT base model and uses much fewer computational resources for training, so we only present the BERT Base performance in the experiment sections. Nevertheless, we still wonder how well BERT large can perform in these unsupervised semantic tasks, so we compare our method with BERT Large in Table~\ref{tb:STS_bert}, Table~\ref{tb:sum_bert}, Table~\ref{tb:phrase_sim_bert}, Table~\ref{tb:phrase_hyper_bert}. As we can see, BERT large is usually better than BERT base in the similarity tasks but performs worse in the hypernym detection task. The BERT's performance gains in similarity tasks might imply that training a larger version of our model might be a promising future direction. 

\subsection{Motivating Examples in Sentence Similarity}
\label{sec:good_example_STS}

In order to further understand when and why our methods perform well, we present some sentences pairs from the MSRvid dataset in STS 2012 in Table~\ref{tb:good_example_STS} and~\ref{tb:good_topics_STS} on which our methods perform well. 

In Table~\ref{tb:good_example_STS}, the first two sentence pairs have relatively high similarities but a lower ratio of overlapping words, so the baseline based on average word embedding (i.e., \textbf{Avg}) underestimates the similarities. Softly removing the stop words (i.e., \textbf{Prob\_avg}) alleviates the problem, but the inverse frequency of words do not completely align with the importance of words in the sentences. 

We visualize our predicted word importance and codebook embeddings in Table~\ref{tb:good_topics_STS}. Combining the estimated word importance with the inverse word frequency (i.e., \textbf{Prob\_avg + Our a}) improves the performance. Finally, computing the similarity between the codebook embeddings (i.e., \textbf{Our c}) leads to the best results. The reason of the improvement might be that the unimportant words in the sentence often do not significantly affect the co-occuring word distribution. Take the second sentence pair as an example, mentioning \emph{``with the big eyes''} does not change the sentence's meaning and facets too much.

On the contrary, the last sentence pair in Table~\ref{tb:good_example_STS} has a low similarity but relatively higher word overlapping. Our model could infer that \emph{riding a horse} is very different from \emph{riding an elephant} because their co-occurring word distributions are different. The appearance of \emph{riding a horse} implies that we are more likely to observe a race topic in nearby sentences, but \emph{riding an elephant} increases the chance of seeing a movie topic instead.



\footnotetext{The number is different from the one reported in \citet{bird-naacl2019} because we use the uncased version (42B), the embedding space our model is trained on, and they use the cased version (840B).}

\subsection{Motivating Examples in Extractive Summarization}
\label{sec:good_example_sum}

In Table~\ref{tb:good_example_sum}, we show the top three sentences that different methods choose to summarize a story about a photographer, Erik Johansson, and his artwork.

In this document, \textbf{Lead-3} does not cover its main points because this article starts with a preamble. Our method selects the first sentence as a good summary because it highlights the main character of the story, Erik Johansson, and his art style. The selected sentences contain the aspects that cover several topics in the whole document.

Average word embedding baselines, \textbf{Sent\_Emb (GloVe)} and \textbf{Sent\_Emb (BERT)}, select the sentences that focus on describing how his artwork is created. Nevertheless, the sentences are hard to understand without the context in the article. We hypothesize that the methods tend to avoid selecting the sentences with diverse aspects because after averaging the word embeddings, the resulting single embedding is not close to the embedding of words in the documents.

Finally, \textbf{W\_Emb (GloVe)} and \textbf{W\_Emb (BERT)} tend to select shorter sentences because we normalize the objective function by the sentence lengths. It is hard to remove the bias of selecting shorter or longer sentences because each sentence is represented by a different number of embeddings.



\section{Experimental Details}
\label{sec:exp_details}


\subsection{Training}
The training algorithm of non-negative sparse coding (NNSC) loss can be seen in Algorithm~\ref{algo:NNSC}. Given the computational resource constraints, we keep our model simple enough to have the  training loss nearly converged after 1 or 2 epoch(s). Since training takes a long time, we do not fine-tune the hyper-parameters in our models. We use a much smaller model than BERT but the architecture details in our transformer and most of its hyper-parameters are the same as those used in BERT. 

The sparsity penalty weights on coefficient matrix $\lambda$ in~\refeqer is set to be 0.4. The maximal sentence size is set to be 50, and we ignore the sentences longer than that. The maximal number of co-occurring words is set to be 30 (after removing the stop words), and we sub-sample the words if there are more words in the previous and next sentence. All words occurring less than 100 times in the training corpus are mapped to <unk>.

The number of dimensions in transformers is set to be 300. For sentence representation, dropout on attention is 0.1. Its number of transformer layers on the decoder side is 5 for $K = 10$, and the number of transformer layers on the decoder side is set to be 1 for $K = 1$ because we do not need to model the dependency of output codebook embeddings. For phrase representation, the number of transformer layers on the decoder side is 2, and the dropout on attention is 0.5. 


All the architecture and hyperparameters (except the number of codebook embeddings) in our models are determined by the validation loss of the self-supervised co-occurring word reconstruction task in \refeqobj. The number of codebook embeddings $K$ is chosen by the performance of training data in each task, but we observe that the performances are usually not sensitive to the numbers as long as $K$ is large enough as shown in our phrase experiments. Furthermore, we suspect that the slight performance drops of models with too large $K$ might just be caused by the fact that larger $K$ needs longer training time and 1 week of training is insufficient to make the model converge.

We use RegexpParser in NLTK~\citep{bird2009natural} to detect the phrase boundary. We use the grammar \textit{NP: {<JJ.*>*<VBG>*<NN.*>+}}. The sentence boundaries are detected using the rule-based pipeline in spaCy\footnote{\url{spacy.io/}} and POS tags are also detected using spaCy.

The lowercased list we use for removing stop words includes \textit{@-@, =, <eos>, <unk>, disambiguation, etc, etc., --, @card@, $\sim$, -, \_, @, \^, \&, *, <, >, (, ), \textbackslash, |, \{, \}, ], [, :, ;, ', ", /, ?, !, ,, ., 't, 'd, 'll, 's, 'm, 've, a, about, above, after, again, against, all, am, an, and, any, are, aren, as, at, be, because, been, before, being, below, between, both, but, by, can, cannot, could, couldn, did, didn, do, does, doesn, doing, don, down, during, each, few, for, from, further, had, hadn, has, hasn, have, haven, having, he, her, here, here, hers, herself, him, himself, his, how, how, i, if, in, into, is, isn, it, it, its, itself, let, me, more, most, mustn, my, myself, no, nor, not, of, off, on, once, only, or, other, ought, our, ours, ourselves, out, over, own, same, she, should, shouldn, so, some, such, than, that, the, their, theirs, them, themselves, then, there, these, they, this, those, through, to, too, under, until, up, very, was, wasn, we, were, weren, what, when, where, which, while, who, whom, why, with, won, would, wouldn, you, your, yours, yourself, yourselves}.

\begin{algorithm*}[!t]
\SetAlgoLined
\SetKwInOut{Input}{Input}
\SetKwInOut{Output}{Output}

\Input{Training corpus, sequence boundaries, and pre-trained word embedding.}
\Output{$F$}
Initialize $F$ \\
\ForEach{$I_t, \bm{W(N_t)}, \bm{W(N_{r_t})}$ \text{in training corpus}}{%
	Run forward pass on encoder and decoder to compute $\bm{F(I_t)}$ \\
	Compute $\bm{M^{O_t}} = \argmin_{\bm{M}} ||\bm{F(I_t)}\bm{M} - \bm{W(N_t)}||^2 + \lambda || \bm{M} ||_1 \forall k,j, \; 0 \leq \bm{M}_{k,j} \leq 1$,  \\
	Compute $\bm{M^{R_t}} = \argmin_{\bm{M}} ||\bm{F(I_t)}\bm{M} - \bm{W(N_{r_t})}||^2 + \lambda || \bm{M} ||_1 \forall k,j, \; 0 \leq \bm{M}_{k,j} \leq 1$, \\
	Run forward pass to compute $L_t$ in \refeqobj \\
	Treat $\bm{M^{O_t}}$ and $\bm{M^{R_t}}$ as constants, update $F$ by backpropagation
}
 \caption{Training using NNSC loss}
 \label{algo:NNSC}
\end{algorithm*}

\subsection{Testing}
The dataset sizes for sentence representation and phrase representation are summarized in Table~\ref{tb:dataset_size_sent} and Table~\ref{tb:dataset_size_phrase}, respectively. In our phrase experiments, we report the test sets of SemEval 2013 and Turney. For Turney dataset, we follow the evaluation setup of~\citet{fct, huang2017learning}, which ignores two unigram candidates being contained in the target phrase, because the original setup~\citep{turney2012} is too difficult for unsupervised methods to get a meaningful score (e.g., the accuracy of \textbf{GloVe Avg} is 0 in the original setting).

For skip-thoughts, the hidden embedding size is set to be 600. To make the comparison fair, we retrain the skip-thoughts in Wikipedia 2016 for 2 weeks.

\section{Randomly Sampled Examples}
\label{sec:more_examples}
We visualize the predicted codebook embeddings and the attention weights computed using~\refeqatt~from 10 randomly selected sentences in our validation set (so most of them are unseen in our training corpus). 

The first line of each example is always the preprocessed input sentence, where <unk> means an out-of-vocabulary placeholder. The attention weights are visualized using a red background. If one word is more likely to be similar to the words in the nearby sentences, it will get more attention and thus highlighted using a darker red color.

The format of visualized embeddings is similar to Table~\reftbvis
. Each row's embedding is visualized by the nearest five neighbors in a GloVe embedding space and their cosine similarities to the codebook embedding.


\colorbox{c87}{Other} \colorbox{c56}{immobilizing} \colorbox{c48}{devices} \colorbox{c57}{such} \colorbox{c60}{as} \colorbox{c62}{a} \colorbox{c73}{Kendrick} \colorbox{c100}{<unk>} \colorbox{c78}{Device} \colorbox{c58}{or} \colorbox{c62}{a} \colorbox{c53}{backboard} \colorbox{c53}{can} \colorbox{c58}{be} \colorbox{c60}{used} \colorbox{c62}{to} \colorbox{c46}{stabilize} \colorbox{c59}{the} \colorbox{c62}{remainder} \colorbox{c64}{of} \colorbox{c59}{the} \colorbox{c32}{spinal} \colorbox{c53}{column} \colorbox{c71}{.} \colorbox{c86}{)} \colorbox{c100}{<eos>} \\
------------------------K=10------------------------ \\
\small
use 0.844 can 0.816 used 0.801  \\
bottom 0.757 front 0.703 sides 0.691  \\
spinal 0.914 nerve 0.739 Spinal 0.700  \\
increases 0.766 decreasing 0.756 increasing 0.749  \\
devices 0.836 device 0.787 wireless 0.703  \\
symptoms 0.726 chronic 0.715 disease 0.692  \\
polymeric 0.674 hydrophilic 0.644 hydrophobic 0.636  \\
backboard 0.899 hoop 0.555 dunks 0.547  \\
column 0.898 columns 0.771 Column 0.584  \\
Kendrick 0.927 Lamar 0.620 Meek 0.611  \\
\normalsize
------------------------K=3------------------------ \\
\small
necessary 0.750 use 0.736 can 0.732  \\
spinal 0.801 thoracic 0.713 nerve 0.702  \\
backboard 0.771 ball 0.629 hoop 0.593  \\
\normalsize
------------------------K=1------------------------ \\
\small
tissue 0.667 prevent 0.659 pressure 0.642  \\
\normalsize

\colorbox{c58}{When} \colorbox{c21}{she} \colorbox{c44}{came} \colorbox{c74}{in} \colorbox{c72}{,} \colorbox{c21}{she} \colorbox{c49}{was} \colorbox{c46}{always} \colorbox{c77}{bound} \colorbox{c71}{to} \colorbox{c65}{be} \colorbox{c51}{loud} \colorbox{c72}{,} \colorbox{c59}{and} \colorbox{c55}{boisterous} \colorbox{c74}{.} \colorbox{c10}{Carrie} \colorbox{c47}{got} \colorbox{c61}{along} \colorbox{c60}{well} \colorbox{c63}{with} \colorbox{c66}{most} \colorbox{c76}{of} \colorbox{c71}{the} \colorbox{c58}{waitresses} \colorbox{c72}{,} \colorbox{c66}{most} \colorbox{c56}{especially} \colorbox{c41}{Vera} \colorbox{c30}{Louise} \colorbox{c60}{Gorman} \colorbox{c85}{-} \colorbox{c60}{Novak} \colorbox{c72}{,} \colorbox{c59}{and} \colorbox{c31}{Alice} \colorbox{c79}{Hyatt} \colorbox{c74}{.} \colorbox{c100}{<eos>} \\
------------------------K=10------------------------ \\
\small
really 0.893 know 0.877 think 0.869  \\
Carrie 0.901 Christina 0.727 Amanda 0.723  \\
daughter 0.816 mother 0.815 wife 0.781  \\
Carrie 0.908 Christina 0.748 Amanda 0.739  \\
came 0.730 went 0.722 had 0.711  \\
Maureen 0.750 Carolyn 0.749 Joanne 0.740  \\
endearing 0.670 downright 0.643 demeanor 0.632  \\
Vera 0.953 Aloe 0.639 vera 0.585  \\
actor 0.672 starring 0.650 comedy 0.639  \\
dancing 0.647 singing 0.598 raucous 0.593  \\
\normalsize
------------------------K=3------------------------ \\
\small
really 0.857 thought 0.853 never 0.846  \\
Carrie 0.881 Amanda 0.801 Rebecca 0.792  \\
waitress 0.596 hostess 0.595 maid 0.591  \\
\normalsize
------------------------K=1------------------------ \\
\small
knew 0.806 she 0.793 thought 0.788  \\
\normalsize

\colorbox{c76}{The} \colorbox{c34}{station} \colorbox{c47}{building} \colorbox{c66}{is} \colorbox{c40}{located} \colorbox{c67}{in} \colorbox{c63}{the} \colorbox{c57}{district} \colorbox{c71}{of} \colorbox{c100}{<unk>} \colorbox{c75}{.} \colorbox{c81}{These} \colorbox{c48}{services} \colorbox{c48}{operate} \colorbox{c70}{on} \colorbox{c63}{the} \colorbox{c58}{Eifel} \colorbox{c35}{Railway} \colorbox{c86}{(} \colorbox{c100}{<unk>} \colorbox{c82}{)} \colorbox{c75}{.} \colorbox{c100}{<eos>} \\
------------------------K=10------------------------ \\
\small
station 0.989 stations 0.847 Station 0.756  \\
Eifel 0.855 Harz 0.673 Cochem 0.642  \\
railway 0.820 railways 0.794 trains 0.782  \\
Germany 0.813 Berlin 0.766 Munich 0.751  \\
north 0.885 south 0.877 east 0.869  \\
building 0.867 buildings 0.794 construction 0.706  \\
located 0.696 operated 0.669 operates 0.668  \\
line 0.877 lines 0.739 Line 0.712  \\
Railway 0.751 Rail 0.622 Railways 0.591  \\
services 0.909 service 0.879 provider 0.708  \\
\normalsize
------------------------K=3------------------------ \\
\small
station 0.918 stations 0.807 railway 0.779  \\
located 0.802 area 0.770 situated 0.734  \\
Aachen 0.706 Eifel 0.701 Freiburg 0.661  \\
\normalsize
------------------------K=1------------------------ \\
\small
station 0.853 railway 0.834 stations 0.748  \\
\normalsize

\colorbox{c76}{Alfred} \colorbox{c100}{<unk>} \colorbox{c100}{<unk>} \colorbox{c77}{(} \colorbox{c68}{born} \colorbox{c53}{January} \colorbox{c71}{22} \colorbox{c79}{,} \colorbox{c61}{1965} \colorbox{c77}{)} \colorbox{c88}{is} \colorbox{c88}{a} \colorbox{c63}{Ghanaian} \colorbox{c66}{businessman} \colorbox{c82}{and} \colorbox{c88}{a} \colorbox{c51}{former} \colorbox{c41}{Honorary} \colorbox{c39}{Vice} \colorbox{c45}{Consul} \colorbox{c79}{of} \colorbox{c51}{Austria} \colorbox{c90}{to} \colorbox{c50}{Ghana} \colorbox{c82}{and} \colorbox{c88}{a} \colorbox{c76}{leading} \colorbox{c59}{member} \colorbox{c79}{of} \colorbox{c79}{the} \colorbox{c49}{National} \colorbox{c45}{Democratic} \colorbox{c49}{Congress} \colorbox{c88}{.} \colorbox{c100}{<eos>} \\
------------------------K=10------------------------ \\
\small
Ghana 0.928 Zambia 0.807 Cameroon 0.796  \\
Committee 0.824 Council 0.740 Commission 0.733  \\
election 0.810 elections 0.788 elected 0.774  \\
Austria 0.923 Germany 0.793 Austrian 0.739  \\
February 0.868 2011 0.865 2012 0.858  \\
Consul 0.898 consul 0.761 consular 0.596  \\
1995 0.970 1994 0.970 1993 0.969  \\
Party 0.886 party 0.707 Parties 0.649  \\
University 0.858 College 0.728 Graduate 0.727  \\
retired 0.575 born 0.572 emeritus 0.543  \\
\normalsize
------------------------K=3------------------------ \\
\small
President 0.778 Chairman 0.765 Committee 0.760  \\
Ghana 0.924 Zambia 0.808 Cameroon 0.806  \\
1999 0.956 1998 0.953 1997 0.951  \\
\normalsize
------------------------K=1------------------------ \\
\small
President 0.723 Affairs 0.664 Minister 0.658  \\
\normalsize

\colorbox{c35}{ISBN} \colorbox{c58}{0} \colorbox{c71}{-} \colorbox{c81}{8063} \colorbox{c71}{-} \colorbox{c66}{1367} \colorbox{c71}{-} \colorbox{c66}{6} \colorbox{c43}{Winthrop} \colorbox{c75}{,} \colorbox{c45}{John} \colorbox{c80}{.} \colorbox{c43}{Winthrop} \colorbox{c74}{'s} \colorbox{c47}{Journal} \colorbox{c75}{,} \colorbox{c30}{History} \colorbox{c79}{of} \colorbox{c44}{New} \colorbox{c47}{England} \colorbox{c58}{1630} \colorbox{c71}{-} \colorbox{c57}{1649} \colorbox{c80}{.} \colorbox{c44}{New} \colorbox{c41}{York} \colorbox{c75}{,} \colorbox{c50}{NY} \colorbox{c66}{:} \colorbox{c44}{Charles} \colorbox{c50}{Scribner} \colorbox{c74}{'s} \colorbox{c57}{Sons} \colorbox{c75}{,} \colorbox{c45}{1908} \colorbox{c80}{.} \colorbox{c100}{<eos>} \\
------------------------K=10------------------------ \\
\small
New 0.936 York 0.915 NY 0.807  \\
J. 0.847 R. 0.810 D. 0.807  \\
ISBN 0.934 Paperback 0.821 Hardcover 0.803  \\
History 0.830 Historical 0.694 War 0.681  \\
0 0.933 1 0.739 3 0.663  \\
Winthrop 0.926 Endicott 0.696 Amherst 0.672  \\
1985 0.840 1981 0.836 1982 0.834  \\
University 0.901 College 0.714 Northwestern 0.692  \\
England 0.875 London 0.718 Britain 0.680  \\
842 0.838 794 0.835 782 0.831  \\
\normalsize
------------------------K=3------------------------ \\
\small
William 0.841 J. 0.789 Robert 0.779  \\
New 0.887 York 0.860 NY 0.727  \\
1626 0.685 1684 0.676 1628 0.675  \\
\normalsize
------------------------K=1------------------------ \\
\small
William 0.686 York 0.676 Charles 0.643  \\
\normalsize

\colorbox{c74}{The} \colorbox{c63}{commune} \colorbox{c81}{is} \colorbox{c79}{represented} \colorbox{c83}{in} \colorbox{c79}{the} \colorbox{c76}{Senate} \colorbox{c74}{by} \colorbox{c62}{Soledad} \colorbox{c64}{Alvear} \colorbox{c76}{(} \colorbox{c82}{PDC} \colorbox{c77}{)} \colorbox{c85}{and} \colorbox{c51}{Pablo} \colorbox{c100}{<unk>} \colorbox{c76}{(} \colorbox{c83}{UDI} \colorbox{c77}{)} \colorbox{c84}{as} \colorbox{c76}{part} \colorbox{c82}{of} \colorbox{c79}{the} \colorbox{c69}{8th} \colorbox{c71}{senatorial} \colorbox{c70}{constituency} \colorbox{c76}{(} \colorbox{c47}{Santiago} \colorbox{c74}{-} \colorbox{c81}{East} \colorbox{c77}{)} \colorbox{c82}{.} \colorbox{c74}{Haha} \colorbox{c60}{Sound} \colorbox{c81}{is} \colorbox{c79}{the} \colorbox{c75}{second} \colorbox{c49}{album} \colorbox{c74}{by} \colorbox{c79}{the} \colorbox{c86}{British} \colorbox{c70}{indie} \colorbox{c82}{electronic} \colorbox{c67}{band} \colorbox{c77}{Broadcast} \colorbox{c82}{.} \colorbox{c100}{<eos>} \\
------------------------K=10------------------------ \\
\small
album 0.919 albums 0.844 songs 0.830  \\
constituency 0.810 election 0.773 elections 0.757  \\
released 0.931 release 0.847 releases 0.783  \\
Santiago 0.859 Juan 0.753 Luis 0.738  \\
Sound 0.890 Audio 0.739 Sounds 0.671  \\
February 0.881 2011 0.878 2010 0.877  \\
Jorge 0.687 Miguel 0.684 Pablo 0.683  \\
Alvear 0.738 Altamirano 0.563 Ruperto 0.552  \\
commune 0.939 communes 0.800 Commune 0.660  \\
\# 0.986 Item 0.523 1 0.507  \\
\normalsize
------------------------K=3------------------------ \\
\small
commune 0.718 communes 0.638 La 0.596  \\
album 0.948 albums 0.865 songs 0.796  \\
election 0.814 elections 0.791 electoral 0.745  \\
\normalsize
------------------------K=1------------------------ \\
\small
album 0.925 albums 0.818 Album 0.745  \\
\normalsize

\colorbox{c70}{As} \colorbox{c72}{a} \colorbox{c71}{result} \colorbox{c78}{,} \colorbox{c64}{the} \colorbox{c51}{Hellfire} \colorbox{c59}{Club} \colorbox{c55}{believed} \colorbox{c61}{that} \colorbox{c65}{it} \colorbox{c63}{would} \colorbox{c67}{be} \colorbox{c75}{in} \colorbox{c61}{their} \colorbox{c74}{best} \colorbox{c72}{interests} \colorbox{c71}{to} \colorbox{c54}{summon} \colorbox{c64}{the} \colorbox{c54}{Phoenix} \colorbox{c71}{and} \colorbox{c84}{merge} \colorbox{c65}{it} \colorbox{c71}{with} \colorbox{c59}{Jean} \colorbox{c53}{Grey} \colorbox{c87}{via} \colorbox{c72}{a} \colorbox{c70}{ritual} \colorbox{c80}{.} \colorbox{c100}{<eos>} \\
------------------------K=10------------------------ \\
\small
want 0.886 way 0.875 sure 0.871  \\
attack 0.722 kill 0.708 enemy 0.708  \\
beings 0.703 manifestation 0.690 spiritual 0.685  \\
disappeared 0.710 apparently 0.681 initially 0.678  \\
Evil 0.719 Darkness 0.694 Demon 0.693  \\
Phoenix 0.950 Tucson 0.688 Tempe 0.658  \\
Hellfire 0.932 hellfire 0.623 Brimstone 0.530  \\
Grey 0.956 Gray 0.788 Blue 0.771  \\
Jean 0.912 Pierre 0.767 Jacques 0.736  \\
Club 0.903 club 0.804 clubs 0.680  \\
\normalsize
------------------------K=3------------------------ \\
\small
even 0.836 somehow 0.823 because 0.816  \\
Hellfire 0.814 Scourge 0.606 Goblin 0.588  \\
Blue 0.738 Grey 0.730 Red 0.693  \\
\normalsize
------------------------K=1------------------------ \\
\small
somehow 0.780 even 0.769 nothing 0.747  \\
\normalsize

\colorbox{c70}{Skeletal} \colorbox{c46}{problems} \colorbox{c78}{,} \colorbox{c22}{infection} \colorbox{c78}{,} \colorbox{c58}{and} \colorbox{c21}{tumors} \colorbox{c51}{can} \colorbox{c63}{also} \colorbox{c42}{affect} \colorbox{c59}{the} \colorbox{c43}{growth} \colorbox{c62}{of} \colorbox{c59}{the} \colorbox{c32}{leg} \colorbox{c78}{,} \colorbox{c51}{sometimes} \colorbox{c59}{giving} \colorbox{c55}{rise} \colorbox{c59}{to} \colorbox{c62}{a} \colorbox{c57}{one} \colorbox{c94}{-} \colorbox{c73}{sided} \colorbox{c51}{bow} \colorbox{c94}{-} \colorbox{c100}{<unk>} \colorbox{c67}{.} \colorbox{c100}{<eos>} \\
------------------------K=10------------------------ \\
\small
symptoms 0.851 disease 0.826 chronic 0.818  \\
often 0.857 usually 0.854 may 0.851  \\
infection 0.849 infections 0.807 infected 0.772  \\
tumors 0.867 tumours 0.834 tumor 0.791  \\
epithelial 0.676 epithelium 0.672 upregulation 0.667  \\
increasing 0.805 increased 0.799 increases 0.798  \\
legs 0.737 shoulders 0.716 fingers 0.715  \\
leg 0.909 legs 0.785 thigh 0.750  \\
mammalian 0.666 gene 0.664 genes 0.654  \\
bow 0.899 bows 0.775 sash 0.574  \\
\normalsize
------------------------K=3------------------------ \\
\small
affect 0.769 significant 0.755 decrease 0.741  \\
disease 0.795 infection 0.782 infections 0.757  \\
epithelial 0.691 extracellular 0.686 epithelium 0.685  \\
\normalsize
------------------------K=1------------------------ \\
\small
disease 0.764 abnormal 0.751 tissue 0.743  \\
\normalsize

\colorbox{c100}{<unk>} \colorbox{c83}{is} \colorbox{c81}{composed} \colorbox{c82}{of} \colorbox{c82}{the} \colorbox{c81}{:} \colorbox{c51}{Jack} \colorbox{c54}{Carty} \colorbox{c94}{may} \colorbox{c89}{refer} \colorbox{c89}{to} \colorbox{c81}{:} \colorbox{c51}{Jack} \colorbox{c54}{Carty} \colorbox{c74}{(} \colorbox{c46}{musician} \colorbox{c75}{)} \colorbox{c74}{(} \colorbox{c46}{born} \colorbox{c50}{1987} \colorbox{c75}{)} \colorbox{c78}{,} \colorbox{c48}{Australian} \colorbox{c46}{musician} \colorbox{c51}{Jack} \colorbox{c54}{Carty} \colorbox{c74}{(} \colorbox{c45}{rugby} \colorbox{c63}{union} \colorbox{c75}{)} \colorbox{c74}{(} \colorbox{c46}{born} \colorbox{c52}{1992} \colorbox{c75}{)} \colorbox{c78}{,} \colorbox{c45}{rugby} \colorbox{c63}{union} \colorbox{c60}{player} \colorbox{c79}{from} \colorbox{c42}{Ireland} \colorbox{c42}{John} \colorbox{c54}{Carty} \colorbox{c74}{(} \colorbox{c73}{disambiguation} \colorbox{c75}{)} \colorbox{c100}{<eos>} \\
------------------------K=10------------------------ \\
\small
February 0.910 April 0.906 June 0.904  \\
States 0.746 American 0.745 United 0.735  \\
Australian 0.848 Zealand 0.749 Australia 0.729  \\
born 0.880 Born 0.686 married 0.596  \\
band 0.598 album 0.596 music 0.583  \\
Carty 0.914 Kelleher 0.621 O'Mahony 0.619  \\
1977 0.967 1978 0.964 1974 0.963  \\
football 0.786 player 0.740 soccer 0.740  \\
union 0.841 unions 0.698 Unions 0.631  \\
Jack 0.800 Jim 0.727 Tom 0.715  \\
\normalsize
------------------------K=3------------------------ \\
\small
1975 0.740 1981 0.738 1978 0.736  \\
Ireland 0.910 Irish 0.782 Dublin 0.747  \\
rugby 0.766 football 0.759 soccer 0.718  \\
\normalsize
------------------------K=1------------------------ \\
\small
England 0.609 Rugby 0.600 Wales 0.596  \\
\normalsize

\colorbox{c73}{He} \colorbox{c54}{designed} \colorbox{c68}{a} \colorbox{c58}{system} \colorbox{c57}{of} \colorbox{c35}{streets} \colorbox{c54}{which} \colorbox{c57}{generally} \colorbox{c69}{followed} \colorbox{c54}{the} \colorbox{c66}{contours} \colorbox{c57}{of} \colorbox{c54}{the} \colorbox{c28}{area} \colorbox{c68}{'s} \colorbox{c53}{topography} \colorbox{c70}{.} \colorbox{c48}{Residential} \colorbox{c27}{neighborhoods} \colorbox{c70}{stretched} \colorbox{c61}{out} \colorbox{c64}{from} \colorbox{c68}{a} \colorbox{c40}{commercial} \colorbox{c56}{and} \colorbox{c46}{service} \colorbox{c84}{-} \colorbox{c49}{sector} \colorbox{c65}{core} \colorbox{c70}{.} \colorbox{c100}{<eos>} \\
------------------------K=10------------------------ \\
\small
creating 0.737 incorporate 0.726 developing 0.720  \\
north 0.861 south 0.859 east 0.855  \\
sector 0.874 sectors 0.790 economic 0.746  \\
neighborhoods 0.840 streets 0.775 neighbourhoods 0.749  \\
City 0.745 Riverside 0.726 Heights 0.709  \\
buildings 0.691 brick 0.672 building 0.658  \\
topography 0.817 vegetation 0.708 forested 0.702  \\
service 0.889 services 0.809 Service 0.663  \\
Building 0.742 Construction 0.613 Project 0.581  \\
1930 0.856 1920s 0.826 1929 0.815  \\
\normalsize
------------------------K=3------------------------ \\
\small
development 0.739 significant 0.720 providing 0.719  \\
streets 0.831 city 0.793 neighborhoods 0.786  \\
County 0.669 Riverside 0.662 City 0.649  \\
\normalsize
------------------------K=1------------------------ \\
\small
buildings 0.765 area 0.761 areas 0.753  \\
\normalsize

\end{document}